%% file: acl_latex.tex
\documentclass[11pt]{article}

\usepackage[final]{acl}

\usepackage{times}
\usepackage{latexsym}
\usepackage[T1]{fontenc}
\usepackage[utf8]{inputenc}

\usepackage{microtype}

\usepackage{inconsolata}

\usepackage{stfloats}
\usepackage{times}
\usepackage{graphicx}
\usepackage{booktabs}
\usepackage{amsmath,amssymb}
\usepackage{multirow}
\usepackage{array}
\usepackage{array}
\usepackage{nccmath}
\usepackage{pifont} % ✓ ✗
\newcommand{\cmark}{\ding{51}} % ✓
\newcommand{\xmark}{\ding{55}} % ✗

\usepackage{tikz}
\usepackage{pgfplots}
\pgfplotsset{compat=1.18}
\usepgfplotslibrary{polar}
\usepackage{pifont}

\usepackage{float}
\usepackage{amsfonts}
\usepackage{listings}
\usepackage{xcolor}
\usepackage{graphicx}
\usepackage{xcolor}

\usepackage{enumitem}
\setlist[itemize]{noitemsep}
\usepackage{booktabs}
\usepackage{multirow}
\usepackage{array}
\usepackage{pifont} % ✓ ✗
\usepackage{colortbl}
\usepackage[most]{tcolorbox}
\usepackage{titlesec}
\usepackage{titletoc}

\lstdefinestyle{promptstyle}{
    basicstyle=\ttfamily\small,
    breaklines=true,
    frame=single,
    backgroundcolor=\color{gray!10},
    columns=fullflexible,
    keepspaces=true
}
\title{Plan First, Judge Later, Run Better: A DMAIC-Inspired Agentic System for Industrial Anomaly Detection}

\author{\textbf{Yongzi Yu}$^{1}$, \ 
\textbf{Ao Li}$^{2}$, \ 
\textbf{Le Wang}$^{3}$, \ 
\textbf{Ziyue Li}$^{4}$, \\
\textbf{Fugee Tsung}$^{2}$, \ 
\textbf{Yuxuan Liang}$^{1}$, \
\textbf{Man Li}$^{\dagger 5}$\\
\textsuperscript{1}The Hong Kong University of Science and Technology (Guangzhou), \\
\textsuperscript{2}The Hong Kong University of Science and Technology, 
\textsuperscript{3}Shanghai University of Finance and Economics, \\
\textsuperscript{4}Technische Universität München, 
\textsuperscript{5}Southwestern University of Finance and Economics\\
\tt\small yyu322@connect.hkust-gz.edu.cn 
\footnotesize{$^{\dagger}$ Corresponding Author}\\
}

\begin{document}
\maketitle
\begin{abstract}
Large language model (LLM) agents have shown promise in automating complex data-analysis workflows, but their reliable deployment remains challenging in high-stakes industrial scenarios. 
Industrial anomaly detection (IAD) is essential for manufacturing quality, safety, and efficiency, yet existing LLM-based IAD agents mainly focus on execution while under-exploiting strategy formulation. 
Consequently, they struggle to handle heterogeneous modalities in a unified and cost-effective manner. 
Inspired by the DMAIC quality-management framework, we propose \textbf{DMAIC-IAD} (DMAIC-inspired Agentic Industrial Anomaly Detection), a \textbf{"Plan First, Judge Later"} multi-agent system that aligns LLM agents with structured industrial problem-solving. 
DMAIC-IAD distills heterogeneous references into standardized operating procedures (SOPs) before strategy generation, and introduces a pre-trained execution-free judge model to rank candidate strategies without costly runtime trials. 
Extensive experiments across four modalities show that DMAIC-IAD improves average detection performance over applicable agentic baselines by \textbf{37.76\%}.
%Large language model (LLM) agents have shown promise in automating complex data-analysis workflows, but reliable deployment remains challenging in high-stakes industrial scenarios.
%Industrial anomaly detection (IAD) is essential for quality, safety, and efficiency in manufacturing and operational systems.
%Most existing LLM-based IAD agents concentrate on the execution level, yet under-exploit the design of strategy formulation. 
%Consequently, they struggle to provide unified handling across heterogeneous modalities, rendering them brittle and costly in real industrial settings.
%Inspired by the well-recognized quality management DMAIC framework, we propose \textbf{DMAIC-IAD} (DMAIC-inspired Agentic Industrial Anomaly Detection), a \textbf{"Plan First, Judge Later"} multi-agent architecture, aligning LLM agents with each phase of industrial problem-solving. 
%A knowledge distillation mechanism is designed prior to the strategy generation to construct standardized operating procedures (SOPs) from heterogeneous references, and a pre-trained, execution-free judge model is further introduced to rank candidate strategies to enable reliable prioritization without costly runtime trials after generation. 
%Extensive experimental results on datasets of various modalities demonstrate the effectiveness of our approach in real-world settings.
%Our code is available in \url{https://anonymous.4open.science/r/DMAIC-IAD-A84E}. 
\end{abstract}

\maketitle
\section{Introduction}
Anomaly detection is the process of identifying deviations from the expected behaviors of normal data \cite{chandola2009anomaly}. In complex industrial systems,  it is widely used to detect faults on the visual surface \cite{cui2023survey, bergmann2019mvtec}, time-series telemetry \cite{hundman2018detecting, wang2024tssurvey}, tabular process \cite{wang2022hybrid, han2022adbench} and graph signals \cite{ding2019deep, liu2022bond}. However, the difficulties of Industrial Anomaly Detection (IAD) lie in the diversity of industrial scenarios \cite{liu2024deep}, where heterogeneous data modalities may occur concurrently within the same scene. For example, a single steel-plate manufacturing line may simultaneously require visual inspection of surface defects, time-series monitoring for equipment health, and detection of communication networks. 

\begin{figure}
  \includegraphics[width=\columnwidth]{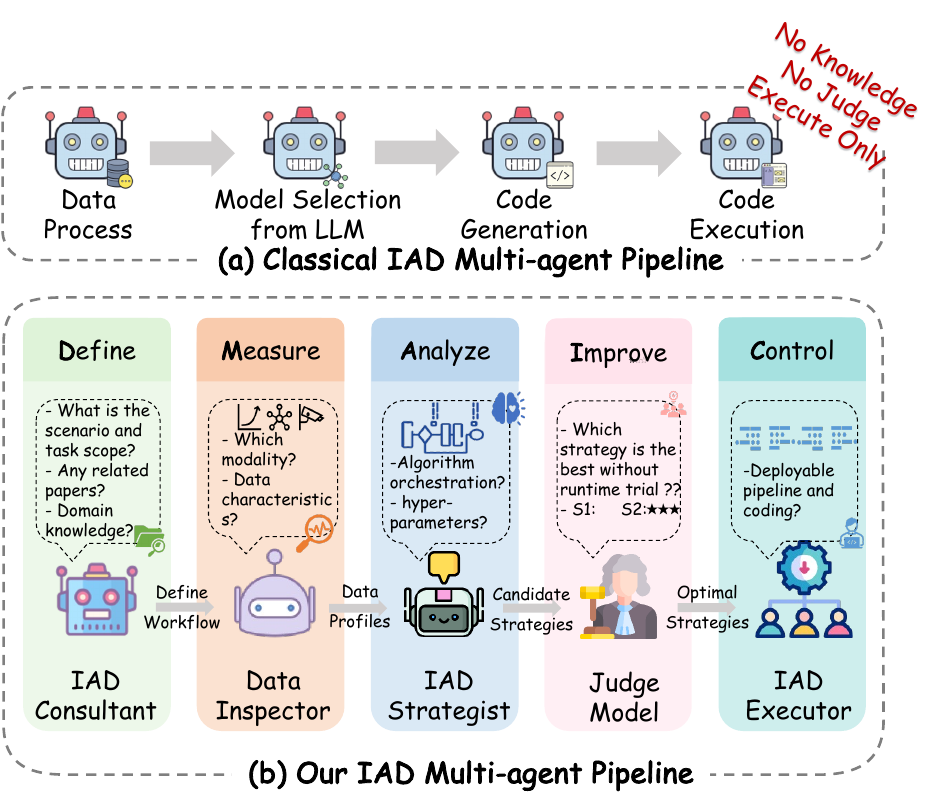}
  \caption{IAD Multi-agent Architecture Comparison: (a) Prior LLM-Generated Strategy Framework vs. (b) DMAIC-Inspired Design Workflow. }
  \label{fig:intro}
  \vspace{-15pt}
\end{figure}

Despite extensive research on modality-specific IAD methods \cite{liu2024deep}, they remain ill-suited for unified handling of heterogeneous multimodal data. Consequently, practical deployment for IAD task still relies on scenario-specific pipeline engineering, and cold-start deployment for novel datasets remains labor-intensive, often requiring manual tuning or costly trial-and-error from scratch due to reliability requirements.

The advent of LLMs has prompted research into agentic systems \cite{li2024survey, guo2024large} for automated anomaly detection \cite{yang2025ad, ji2025autoiad}. Summarized in Fig. \ref{fig:intro}, existing pipelines primarily couple heuristic data processing with direct code generation. Critically, these methods disproportionately emphasize the execution stage while under-exploiting high-level strategy formalization. This imbalance leads to significant reliability risks, as generated strategies cannot be assessed without incurring actual deployment costs. While iterative refinement via RL \cite{hachaj2025explainability, liang2025review} or self-evaluation \cite{fang2025comprehensive} offers a potential remedy, the associated computational burden renders it impractical for real-time industrial needs. Consequently, a gap remains in building \textbf{reliable LLM-based agents} that can achieve truly dependable and autonomous IAD workflows.

% Related IAD agentic Work & Persistent Weakness
%With the recent advent of LLMs, a few studies have explored agentic, automated strategy generation for anomaly detection \cite{yang2025ad, ji2025autoiad}. These frameworks can be summarized as Fig. \ref{fig:intro} (a), which is composed of heuristic data processing, strategy selection, and LLM-based code generation. They mainly concentrate on the execution stage (code generation), ignoring the importance of decision making and strategy formalization. Therefore, they suffer from critical reliability limitations and cannot validate generated strategies before incurring deployment costs. Despite in the common area, LLM self-evaluation and Reinforcement Learning (RL) have been implemented for strategy selection and optimization, but it requires high cost of computation and time. Consequently, they fall short of enabling dependable and fully automated IAD pipelines in industrial settings. 

% Introducing DMAIC as an Organizing Skeleton
To introduce discipline and process rigor to IAD automation, we draw inspiration from DMAIC (Define-Measure-Analyze-Improve-Control), which is a data-driven process improvement framework with deep industrial roots in manufacturing, supply chain management, and service industries \cite{de2012analysis}. It provides a logical, staged roadmap that systematically defines problems, measures current states, analyzes data, improves strategies, and implements controls to sustain results \cite{ishak2019quality, smketkowska2018using}, which closely parallel our industrial anomaly detection task. Therefore, we adopt DMAIC as an explicit organizational framework (Fig. \ref{fig:intro}) to impose discipline, auditability, and stagewise agent responsibilities for \textbf{strategy formulation}, \textbf{pre-execution evaluation}, and \textbf{deployment}.

% Proposed Framework Overview
Building upon this DMAIC-inspired perspective, we propose \textbf{DMAIC-IAD} (DMAIC-inspired Agentic Industrial Anomaly Detection), a multi-agent system designed for the practical constraints of IAD. 
This system first distills domain references into scenario-specific standardized operating procedures (SOPs) to support unified handling of diverse scenarios and address cold-start circumstances. Besides, we construct dataset-specific profiles from dataset samples or descriptions. These artifacts ground downstream strategy generation and evaluation.
Then we build a pre-trained execution-free judge model to automatically score candidate strategies generated from LLMs without running them, thereby avoiding costly trials. Finally, an executor compiles the selected strategy into runnable workflows and produces auditable anomaly reports. By decoupling planning from execution, our system enables efficient, pre-execution assessment of strategy quality, and thereby realizes the principle "\textbf{plan first, judge later, run better}".

 % Contributions
The main contributions of this work are summarized as follows:
\begin{itemize}[leftmargin=*]
  \item We propose a \textbf{DMAIC-inspired agentic framework} for automated IAD to impose discipline, auditability, and stagewise agent responsibilities, which integrates domain knowledge, dataset profiling, and strategy evaluation into a structured pipeline.
  \item We introduce the \textbf{knowledge distillation mechanism} that standardizes SOP construction from heterogeneous references, improving cold-start strategy quality for new scenarios.
  \item We further design an \textbf{execution-free judge model} to predict plan–task compatibility from structural and contextual signals, enabling reliable prioritization of candidate strategies without costly runtime trials.
  \item Systematic experiment  demonstrate that in the majority of cases, the performance of our DMAIC-inspired framework is superior to current state-of-the-art methods with average improvement of \textbf{+37.76\%}.

\end{itemize}

\section{Related Work}
%非常相关的两篇文献：ADAGENT, AutoIAD

\paragraph{Industrial Anomaly Detection.}
%缩短，原版放到Tab.tex的最后面了
%加了一个总结句，把原本的两段Reconstruction-based methods和Feature embedding-based methods合并了
Industrial anomaly detection has been extensively studied under unsupervised settings, where only normal data are used for training \cite{chalapathy2019deep, ruff2021unifying}. Existing methods mainly include reconstruction-based models, feature embedding-based methods \cite{roth2022towards, kirichenko2020normalizing, xuanomaly, su2019robust, xu2018unsupervised}, as well as forecasting- and graph-based approaches for modeling temporal and relational dependencies \cite{sakurada2014anomaly, park2018multimodal, wyatt2022anoddpm, hundman2018detecting, lv2023adaptive}. Most deep learning-based methods remain model-centric and scenario-dependent, requiring task-specific design or retraining and thus limiting scalability and generalization across heterogeneous industrial environments \cite{chalapathy2019deep, li2025survey}.

\begin{figure*}[t]
  \includegraphics[width=\textwidth]{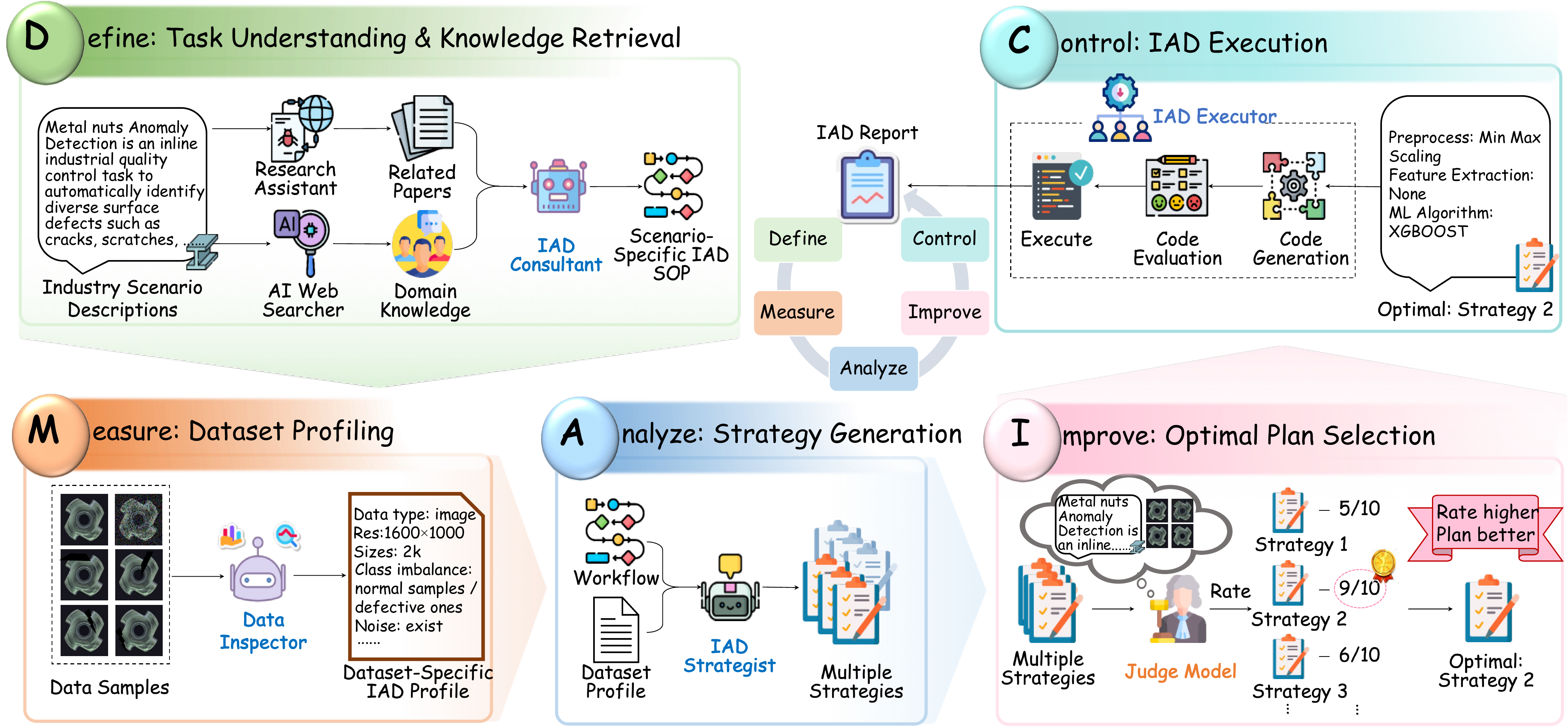}
  \caption{Overall Framework of DMAIC-IAD: Given scenario descriptions and data samples, it distills domain references into \textbf{scenario-specific SOPs}, and constructs \textbf{dataset-specific profiles} (Define-Measure). Using these priors, it generates candidate strategies, followed by a \textbf{pre-trained execution-free judge model} to score without runtime trials (Analyze–Improve). The selected strategy is compiled into a runnable workflow and produces auditable anomaly reports during Control, closing the DMAIC loop.
  %(b) The pretrained judge model in the "Improve" phase. The judge ingests structured strategy specifications together with scenario SOPs and dataset profiles, encodes structural and contextual cues, and outputs calibrated plan–task compatibility scores at inference time to enable pre-execution ranking and safe pruning of LLM-generated strategies.
  }
  % The input of this system is the information description and dataset. Dataset will go through the pre-analysis agent to find out the data structure and status. Text description will produce the key words and request related knowledge to the SOP agent. SOP, data structure and description will be input into planner. The planner will generate several plans. Then judge model will evaluate these plans. We execute the highest plan in the execution block.
  \label{fig:structure}
  \vspace{-8pt}
\end{figure*}
\paragraph{Agentic Anomaly Detection System.}
Early agent-based anomaly detection methods focused on distributed monitoring and were later extended to collaborative multi-agent systems for improved robustness and scalability \cite{mcarthur2005agent, garcia2019multi}. Recent large model-empowered agents have been explored for industrial anomaly detection across time-series \cite{gu2025argos, yang2025agent, qin2025mas}, visual \cite{miao2025agentiad}, and log-based scenarios \cite{ji2025lemad, harbola2025prescriptive}. However, most of these methods use large models as fixed processors rather than adaptive decision-makers for heterogeneous industrial scenarios. Recent frameworks such as AD-AGENT \cite{yang2025ad} and AutoIAD \cite{ji2025autoiad} automate executable pipeline construction but mainly focus on implementation-level optimization, with limited attention to high-level decision-making over model selection and scenario adaptation. More detailed related work is provided in Appendix \ref{appendix:related work}.

% However, most existing LLM-based multi-agent systems primarily emphasize code or action generation under predefined operational procedures. In contrast, our work shifts the focus toward strategic planning by integrating experience knowledge and a dedicated plan selection mechanism, enabling more efficient and adaptive multi-agent collaboration across heterogeneous industrial environments.

\section{Methodology}

\subsection{Architecture of DMAIC-IAD}

Our DMAIC-IAD framework adopts a process-oriented formalization to explicitly encode agent roles, primary artifacts, and operational mappings onto DMAIC phases, as presented in Fig. \ref{fig:structure}. Concretely, the system is expressed by the tuple $\Omega = \langle \mathcal{R}, \mathcal{D}, \mathcal{P}, \mathcal{K}, \Pi \rangle$, where 
$\mathcal{R}$ denotes the domain of task and scenario descriptions, 
$\mathcal{D}$ represents the dataset space, 
$\mathcal{P}=\{\text{Define, Measure, Analyze, Improve, Control}\}$ encompasses the ordered DMAIC phases and the artifacts they produce, 
$\mathcal{K}$ signifies the external knowledge base (integrating both static literature and real-time web search), 
and $\Pi$ is the set of executable strategies.
The framework maps inputs $(r,d)\in\mathcal{R}\times\mathcal{D}$ through the ordered DMAIC phases $\mathcal{P}$ to select and execute a strategy $\pi\in\Pi$, guided by $\mathcal{K}$. There is a \textbf{case study} in Appendix \ref{case study}, and a notation summary of this work is in Appendix \ref{Notation}.

% We formalize the \textsc{IAD} Agentic Framework as a hierarchical state-space system designed to automate the industrial anomaly detection lifecycle. Drawing inspiration from the standard DMAIC management protocol, the system is defined by the tuple $\Omega = \langle \mathcal{R}, \mathcal{D}, \mathcal{S}, \mathcal{K}, \Pi \rangle$. Here, $\mathcal{R}$ denotes the domain of task descriptions, $\mathcal{D}$ represents the dataset space, $\mathcal{S}$ encompasses the dynamic reasoning state, $\mathcal{K}$ signifies the external knowledge base (integrating both static literature and real-time web search), and $\Pi$ is the set of executable strategies. The framework operates by mapping inputs from $\mathcal{R} \times \mathcal{D}$ through a sequential optimization process governed by five distinct phases.

The workflow is operationalized by specialized agents that act as transformation operators within the DMAIC cycle:
\begin{itemize}[leftmargin=*]
\item \textbf{Define:} The \textit{IAD Consultant Agent} initiates the process by interpreting the industry scenario ($\mathcal{R}$) and distilling domain-specific knowledge from $\mathcal{K}$ into a canonical SOP schema. The resulting \emph{scenario-specific SOP} defines the problem scope and operational constraints for downstream phases.
\item \textbf{Measure:} The \textit{Data Inspector Agent} conducts rigorous data analysis on $\mathcal{D}$, identifying distribution shifts and structural requirements to ensure data readiness.
\item \textbf{Analyze:} Synthesizing outputs from the previous phases, the \textit{IAD Strategist Agent} formulates a set of candidate strategies $\Pi$, mapping the problem space to potential algorithmic solutions.
\item \textbf{Improve:} To optimize the strategy selection, a \textit{Judge Model} evaluates the proposals from the Strategist, scoring them to identify the optimal plan for execution.
\item \textbf{Control:} Finally, the \textit{IAD Executor Agent} implements the strategy through code generation, validation, and execution, and gives a final report.
\end{itemize}

% \subsection{Task Understanding and Knowledge \re{Distillation}}
\subsection{Define: Understanding and Knowledge Distillation}
\label{define}

The initialized \textit{Define} phase constitutes the cognitive anchor of the system, executed by the \textit{Consultant Agent}. This agent orchestrates a dual-channel retrieval mechanism comprising a Research Assistant for academic literature and an AI Web Searcher for real-time online resources. We formalize this retrieval-and-distillation process as $\Psi_{cons}: \mathcal{R} \times \mathcal{K} \to \mathcal{W}$. Given an unstructured scenario description $r \in \mathcal{R}$, the agent first extracts semantic keywords to query the external knowledge base $\mathcal{K}$. Subsequently, the LLMs synthesizes the aggregated multi-source context into a structured SOP, denoted as $w$.
\begin{equation*}
w = \Psi_{cons}(r, \mathcal{K}) = \{d_{scen}, \vec{o}_{steps}, \mathcal{M}_{rec}\},
\end{equation*}
where $d_{scen}$ provides a refined definition of the anomaly detection scenario, $\vec{o}_{steps}$ represents the ordered sequence of methodological steps, and $\mathcal{M}_{rec}$ denotes the set of recommended models derived from the retrieval results. A specific SOP case is provided in Appendix Fig. \ref{fig:sop_tile}.

\subsection{Measure: Dataset Profiling}

The \textit{Measure} phase is governed by the \textit{Data Inspector Agent}, which grounds the abstract SOP in the empirical reality of the input data. We formalize this process as a structured diagnostic function $\Psi_{insp}: \mathcal{D} \to \mathcal{M}$. To maintain computational efficiency while ensuring context awareness, the agent operates on a representative subset of samples extracted from the raw dataset $D$, augmented by pre-computed metadata such as label cardinality. The LLM processes these inputs to generate a comprehensive data profile $\mathbf{m} \in \mathcal{M}$:
\begin{equation*}
    \mathbf{m} = \Psi_{insp}(\{x_i\}_{i=1}^k, |Y|) = \langle \phi_{feat}, \tau_{task}, \delta_{miss} \rangle, 
\end{equation*}
where $\{x_i\}_{i=1}^k$ denotes the few-shot data samples and $|Y|$ represents the label count. The resulting output tuple encodes the intrinsic data characteristics ($\phi_{feat}$), the inferred task category ($\tau_{task}$), and a data integrity assessment ($\delta_{miss}$), specifically highlighting issues such as missing values. This structured analysis combines with the SOP to form the context state, ensuring that subsequent planning is not only theoretically sound but also strictly compatible with the dataset's physical constraints. The demo of one dataset profile is in Appendix Fig. \ref{fig:data_profile_tile}.

\subsection{Analyze: Strategy Generation}

In the \textit{Analyze} phase, the \textit{Strategist Agent} transitions the system from passive observation to active architectural planning. We model this transformation via the generative function $\Psi_{strat}: \mathcal{W} \times \mathcal{M} \to \Pi^K$.

Conditioned on the semantic constraints of the SOP $w$ and the empirical characteristics from the data profile $\mathbf{m}$, the LLM synthesizes a diverse set of $K$ candidate strategies. We formalize this generation process as:
\begin{equation*}
\Pi^K = \Psi_{strat}(w, \mathbf{m}) = \{ \pi_k \}_{k=1}^K,
\end{equation*}
where $\pi_k = \{ \tau_j^{(k)} \}_{j=1}^{L_k}$.

Crucially, each strategy $\pi_k$ extends beyond a high-level DAG topology, and it is instantiated as a granular, structured execution sequence of length $L_k$. There are three kinds of steps in each strategy: data preprocessing, feature extraction, and anomaly detection method. Each step is defined by a tuple $\tau_j = \langle i, \alpha_{act}, \mu_{algo}, \theta_{param} \rangle$, which explicitly mandates the step index ($i$), the required action type ($\alpha_{act}$), the specific algorithmic method ($\mu_{algo}$), and its corresponding hyperparameters ($\theta_{param}$). There is a strategy case in Appendix Fig. \ref{fig:plan_example_tile}. This rigorous specification ensures the system explores the solution space with executable precision rather than converging on ambiguous abstract workflows.

\subsection{Improve: Execution-Free Judge Model}

The \textit{Improve} phase addresses the critical "Plan Selection Problem" by identifying the optimal strategy $\pi^*$ from the candidate set $\Pi^K$. A naive approach might employ LLMs for self-evaluation \cite{saha2025learning}; however, the candidate strategies are natively generated by an LLM (the Strategist), and a general LLM lacks the empirical grounding required to distinguish between linguistically plausible and practically executable plans. Furthermore, while RL offers a theoretical pathway for policy optimization \cite{estornell2025train}, the computational overhead of iterative gradient updates on large-scale models proves prohibitive for agile industrial deployment. Consequently, we propose a cost-effective, lightweight \textit{Judge Model} that acts as a surrogate for physical execution, distilling historical experience into a predictive scoring mechanism.

We formalize the Judge as a learnable utility function $\Phi_{judge}: \Pi \times \mathcal{R} \to \mathbb{R}$, parameterized by $\theta$. The architecture leverages a semantic alignment framework: for each candidate strategy $\pi_k$ and the scenario context $r$, we employ a \textbf{pre-trained Sentence Transformer} $E(\cdot)$ to map the textual constraints into a dense latent space. These embeddings are concatenated and fed into a Multilayer Perceptron (\textbf{MLP}) regressor. Crucially, the model is trained via supervised learning to predict a composite utility score $\hat{y}_k$, which approximates the weighted performance metrics observed in historical execution logs:
\vspace{-0.5em}
\begin{equation*}
\hat{y}_k = \mathcal{F}_{mlp}\left( E(\pi_k) \oplus E(r) ; \theta \right) \approx \sum_{j=1}^{M} \lambda_j \cdot \mu_j^{(hist)},
\end{equation*}
\vspace{-0.2em}
where $\mu_j^{(hist)}$ represents the ground-truth historical metrics associated with similar strategy-scenario pairs, and $\lambda_j$ denotes the importance weight of the $j$-th metric. By freezing the encoder and training only the lightweight MLP head, this design decouples high-cost reasoning from high-frequency evaluation. The system finally performs deterministic maximization, $\pi^* = \operatorname*{arg\,max}_{\pi_k} \hat{y}_k$, ensuring the selected plan is grounded in data-driven empirical evidence rather than hallucinated confidence.

\subsection{Control: IAD Execution}

The final \textit{Control} phase operationalizes the theoretically optimal plan $\pi^*$ into a concrete, verifiable software artifact. This process is orchestrated by the \textit{Executor Agent} via a streamlined code generation architecture. We model this as a state-refinement loop governed by a generative coding function $\Psi_{code}: \Pi \to \mathcal{C}$ and a runtime validation oracle $V$.

The agent initiates the process by instantiating the plan into a raw script $c^{(0)} = \Psi_{code}(\pi^*)$. The oracle $V$ subsequently attempts execution, checking for syntax compliance, API alignment, and runtime integrity. If faults are detected, a diagnostic feedback trace $\epsilon^{(t)}$ triggers a self-correction step:
\vspace{-0.2em}
\begin{equation*}
c^{(t+1)} = \Psi_{refine}(c^{(t)}, \epsilon^{(t)}) \quad \text{for} \quad t < T_{max}
\end{equation*}
\vspace{-0.2em}
This iterative cycle continues until validation success or the iteration cap $T_{max}$ is exhausted. Upon successful deployment, the system computes the final performance metrics and synthesizes a human-readable \textit{IAD Report}, encapsulating both the quantitative detection results and the qualitative methodological justification.

\subsection{Efficient Experience Reuse}

To obviate redundant computation for recurring scenarios, the framework integrates a lightweight, hash-based \textbf{memory mechanism}. We operationalize this as a dual-key lookup table mapping input fingerprints to intermediate artifacts: $\mathcal{M}=\{H(r)\mapsto w,\; H(D)\mapsto \mathbf{m}\}$. Here, $H(\cdot)$ represents a deterministic hashing function applied to the raw task description $r$ or the dataset schema $D$.

Upon initialization, the system executes a lookup-first strategy. If a fingerprint match is detected for the task fingerprint $H(r)$, the system retrieves the proven SOP $w$, bypassing the Consultant Agent's synthesis overhead. Similarly, a match for $H(D)$ allows the immediate reuse of the data profile $\mathbf{m}$. This modular design supports partial caching, which could effectively apply established workflows to novel data instances ($r_{match}, D_{new}$). Under experience reuse, the lookup operation achieves average-case $O(1)$ complexity. Detailed description is provided in Appendix \ref{app:experience_reuse}.

\section{Experiment}
In the experimental section, we compare the anomaly detection performance of DMAIC-IAD with baseline methods and assess its computational cost relative to previous approaches. Besides, we also perform the ablation study examining the system's performance in the absence of the judge model or the SOP in \ref{Performance} and \ref{Efficiency}. In \ref{judge model analysis}, we visualize the selection performance of the judge model. And we investigate the impact of utilizing different LLMs as the strategist in \ref{strategist}. Furthermore, we conduct the comparison with specific-modality models in Appendix \ref{app:modality_specific_comparison}, a parameter experiment in Appendix \ref{parameter experiment}, and a case study in Appendix \ref{case study}.

\subsection{Experiment Setting}
\paragraph{Data Description.}
In our experiment, we utilize datasets across four distinct modalities: numeric, time series, graph, and image. We select two representative datasets for each modality. Specifically, the numeric category includes vertebral and arrhythmia datasets from the ADBench \cite{han2022adbench} classical collection. For time series analysis, we utilize PSM and SWaT datasets from Time Series Library \cite{wang2024tssurvey} to validate. The graph datasets are books and enron from BOND \cite{liu2022bond}, and the image modality comprises metalnut and tile datasets in Benchmark MVTecAD \cite{bergmann2019mvtec}. 
\input{tables/main_experiment}
\paragraph{Evaluation Metrics.}
Given the severe label imbalance prevalent in individual datasets, we primarily employ AUROC (Area Under the Receiver Operating Characteristic Curve) and AUPRC (Area Under the Precision-Recall Curve) to evaluate the performance of the generated code \cite{alzarooni2025anomaly}. The specific description of AUROC and AUPRC is in Appendix \ref{auroc}. To provide a comprehensive system evaluation, we also incorporate Success Rate (SR), execution time, Prompt Token (PT), and Completion Token (CT). SR defines the proportion of code instances that execute successfully. Execution time measures the latency from system initialization to the final code generation. PT refers to the volume of inputs processed by the LLMs, and CT aggregates the sum of both input and output tokens across all interactions.

\paragraph{Agent Model Setting.}
The agents within our system utilize distinct backend models selected according to their functional requirements. Specifically, the Consultant and Data Inspector employ GPT-4o \cite{achiam2023gpt}. The Strategist Agent operates on GPT-5-Mini \cite{singh2025openai}, while the Code Generator leverages Claude-Sonnet-4.5. In Exp. \ref{strategist}, we further investigate the impact of model capability on system performance by comparing GPT-3.5-Turbo \cite{brown2020language}, GPT-4o-Mini \cite{achiam2023gpt}, and GPT-5-Mini as alternative backends for the Strategist Agent. Detailed configuration is provided in Appendix \ref{sec:llm-config}.

\paragraph{Judge Model Setting.}
For training of the Judge Model, we selected a representative dataset for each modality to form the training set. This included the optdigits dataset for tabular data, the MSL dataset for time series data, the weibo dataset for graph data, and the screw dataset for image data. For each of these datasets, we utilized the Strategist to generate and execute 50 strategies, resulting in 200 strategy-metrics pairs. The final Judge Model was trained on these 200 strategy-metrics pairs aggregated from the four datasets. The training datasets are independent of the datasets in the experiment. More details are in Appendix \ref{Judge Model Detail}.

%Specifically, the input of the judge model is the keywords extracted from the scenario description and strategies, including the dataset name, datatype, task type, data preprocessing method, feature extraction method, and anomaly detection method. 

\paragraph{Baseline Methods.}
In our comparative analysis, we evaluate the proposed system against three baseline methods: AD-AGENT, AutoIAD, and a "Strategist only" variant. Specifically, AD-AGENT employs LLMs to select algorithms from established repositories for time series, tabular, and graph modalities, whereas AutoIAD is tailored to the image domain for model selection and design, shown in the modality part of Tab. \ref{tab:modality_benchmark_comparison}. The "Strategist only" baseline serves as an ablation study of our framework. It retains only the Analyze and Control (AC) components while omitting the Define, Measure, and Improve (DMI) stages.

\subsection{IAD Performance Comparison}
\label{Performance}
To validate the effectiveness of our method in practical IAD scenarios, we compare different frameworks using AUROC and AUPRC across eight datasets, as shown in Tab.~\ref{tab:main}.

\textbf{DMAIC-IAD Achieves Strong Overall Performance Across Modalities.}
As shown in Tab.~\ref{tab:main} and Tab.~\ref{tab:ours_vs_modality_specific}, our method achieves superior performance on most datasets. It improves AUROC over AD-AGENT by \textbf{+37.0\%} on vertebral, \textbf{+44.8\%} on SWaT, and \textbf{+86.4\%} on enron. For image data, it also surpasses AutoIAD by \textbf{+9.1\%} on metalnut. Overall, our method outperforms AD-AGENT and AutoIAD by \textbf{37.76\%} on average, demonstrating strong robustness across heterogeneous modalities.

\textbf{Strategist Only is not Enough.}
The \textit{Strategist only} baseline shows unstable performance and is clearly inferior to the full DMAIC framework, indicating that the simple AC architecture is insufficient. For example, its AUROC is only 0.6190 on vertebral, compared with 0.9617 for our method, yielding a \textbf{+55.4\%} improvement. The gap is larger on SWaT, where our method improves AUROC by 118.7\%, and on enron, where AUROC increases from 0.1160 to 0.9320. These results validate the necessity of the full DMAIC design.

\textbf{The Judge Model Effectively Selects High-Performing Strategies.}
The Judge Model further improves strategy selection by filtering suboptimal candidates, as later analyzed in Section~\ref{judge model analysis}.

\textbf{SOPs Substantially Enhance System Performance.}
The ablation results in Tab.~\ref{tab:main} show that removing SOPs causes consistent performance degradation. Without SOPs, AUROC drops by \textbf{17.2\%} on vertebral, AUPRC also drops sharply on SWaT and enron, although the absolute AUPRC on enron remains small. These results indicate that SOPs are not merely auxiliary guidance, but a foundational component for stable and high-performing IAD.

\subsection{Efficiency Analysis}
\label{Efficiency}

This experiment investigates the performance of various methods in terms of success rate, system execution time, and token consumption. The results in Tab.~\ref{tab:success_rate} show that our method achieves the highest success rate, and a deeper analysis reveals the following insights into the trade-offs between cost and performance.

\textbf{Our Method Balances Performance and Cost.}
As shown in Tab.~\ref{tab:success_rate}, our method achieves the highest success rate of \textbf{78.19\%}, outperforming AutoIAD by \textbf{3.83\%}. Although it takes the longest execution time, its token consumption is only 20,467, which is \textbf{78.89\%} lower than AutoIAD's 96,906. This shows that our framework achieves strong performance with much lower token cost, making it attractive when detection quality and token cost are prioritized over latency.

\textbf{Latency Comes from Reliability-Oriented Design.}
The longer execution time of our method mainly results from two reliability-oriented components. First, our framework generates multiple candidate strategies before Judge Model selection. Comparing DMAIC-IAD with wo Judge Model in Tab.~\ref{tab:success_rate}, the execution time decreases by about 144.60s, which is mainly introduced by repeated strategy generation for constructing the candidate pool. Second, the Control phase adopts an execution-validation-refinement loop. This process improves execution robustness but increases wall-clock time. Therefore, the latency of our method is mainly the cost of more reliable strategy selection and code execution, while its token consumption remains much lower than AutoIAD.
\input{tables/success_rate_tabel}

\textbf{SOPs are Essential for Reliable Performance.}
Removing SOPs causes substantial degradation across metrics. The success rate drops from \textbf{78.19\%} to \textbf{69.33\%}, and the AUROC/AUPRC scores also decline notably, as shown in Tab.~\ref{tab:main}. For example, on SWaT, the AUPRC drops by over \textbf{91\%}. Although SOPs increase execution time, their removal severely harms both task success and detection quality, confirming their foundational role in the framework.

\subsection{Judge Model Discussion}
\label{judge model analysis}
\begin{figure}[ht]
    \centering

        \includegraphics[width= \columnwidth]{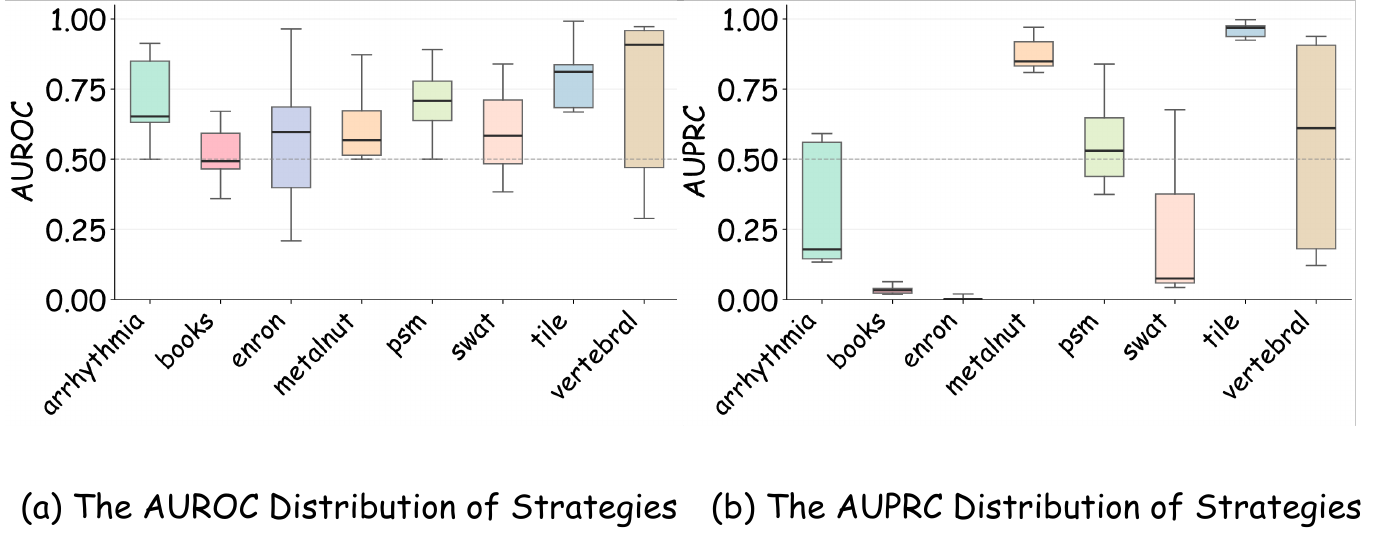}
        \caption{Performance Distribution Box Plot of Strategies without Judge Model: (a) shows the AUROC distribution and (b) shows the AUPRC distribution.}
        \label{fig:box}
        \vspace{-10pt}
\end{figure}
\begin{figure}[ht]
    \centering
        \includegraphics[width=\columnwidth]{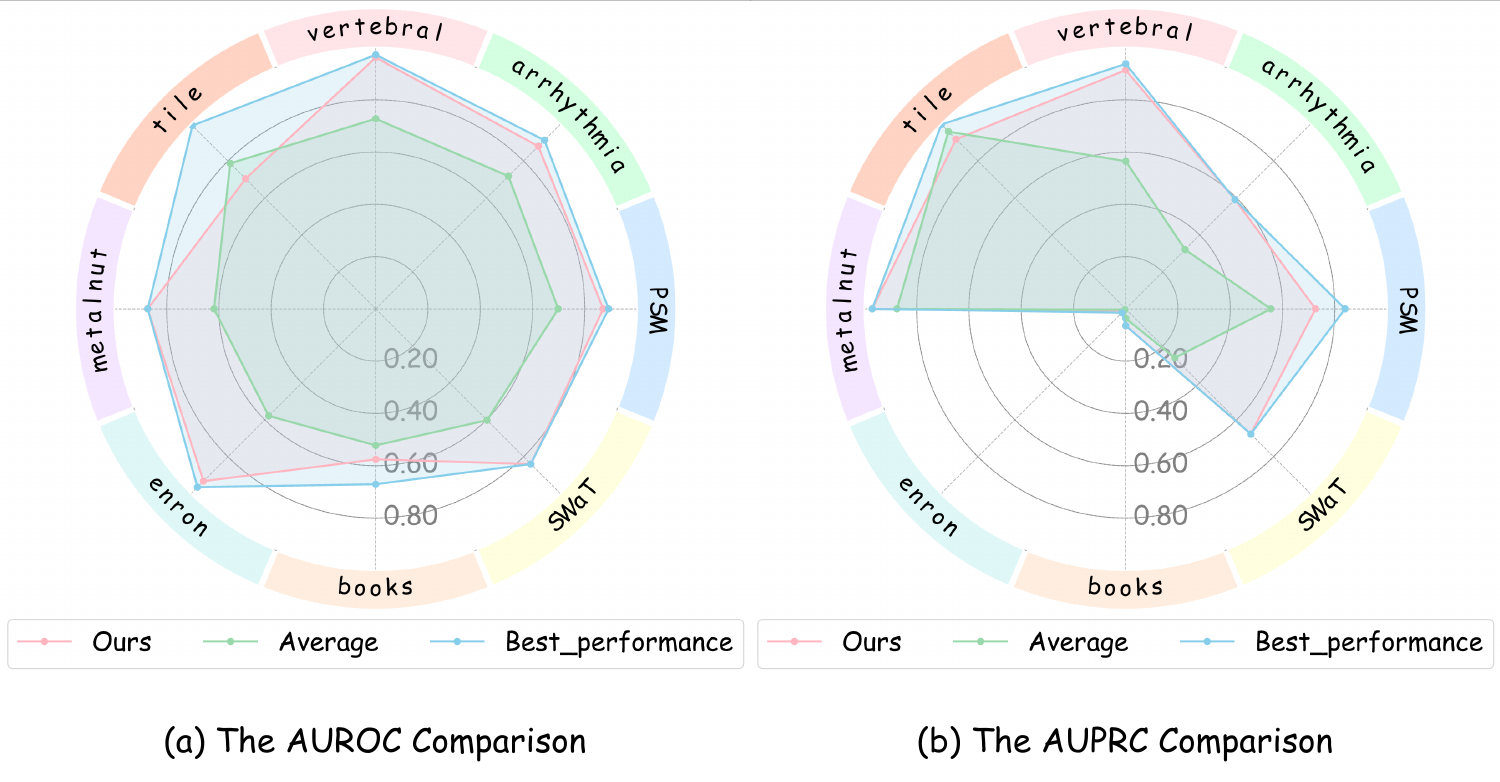}
        \caption{Comparison of Performance between Ours and Strategies: Average means the average performance without judge model and Best\_performance means the best performance without judge model. (a) shows the AUROC comparison on eight datasets and (b) shows the AUPRC metric.}
        \label{fig:judge model}
        \vspace{-15pt}
\end{figure}

%我们将strategist生成的strategies进行执行，对生成的结果找最大值作为Best_performance。对生成的结果求平均作为fig3中的Average。用judge model后的结果和直接生成的strategies做对比。
The Judge Model is crucial for selecting superior strategies from a highly variable candidate pool. As shown in Fig. \ref{fig:judge model}, candidate strategies represented by \textit{Ours wo Judge Model} often contain many suboptimal solutions, making robust pre-execution selection necessary.

\textbf{The Judge Model Consistently Improves Strategy Selection.}
Guided by the Judge Model, our method consistently outperforms the average candidate strategy. As shown in Fig.~\ref{fig:box}, candidate AUROC ranges from 0.4998 to 0.9128 on arrhythmia, while our Judge-guided method achieves 0.8799, \textbf{22.5\%} above the average. And from 0.5790 to 0.9320 on enron, yielding a \textbf{61.0\%} improvement, as shown in Tab. \ref{tab:main} and Fig. \ref{fig:judge model}. It also improves AUPRC by \textbf{30.7\%} on PSM and reaches the best observed performance on several datasets, such as arrhythmia, SWaT, and metalnut. These results show that the Judge Model effectively filters suboptimal strategies and pushes performance toward the upper bound of the candidate pool.

\textbf{Limitations of the Judge Model.}
The Judge Model can still fail on out-of-distribution scenarios within a known modality. As shown in Fig. \ref{fig:judge model}, on the tile dataset, an out-of-distribution texture-based image category, it selects a strategy with an AUROC of 0.7043. This is likely because we conjecture that weak and strong strategy combinations may become similar in the learned feature space, misleading the model's selection. This limitation suggests that the Judge Model requires stronger scenario- and dataset-aware generalization.

\subsection{Analysis of Different LLM-based Strategist}
\label{strategist}
\begin{figure}
\centering
  \includegraphics[width=0.9\columnwidth]{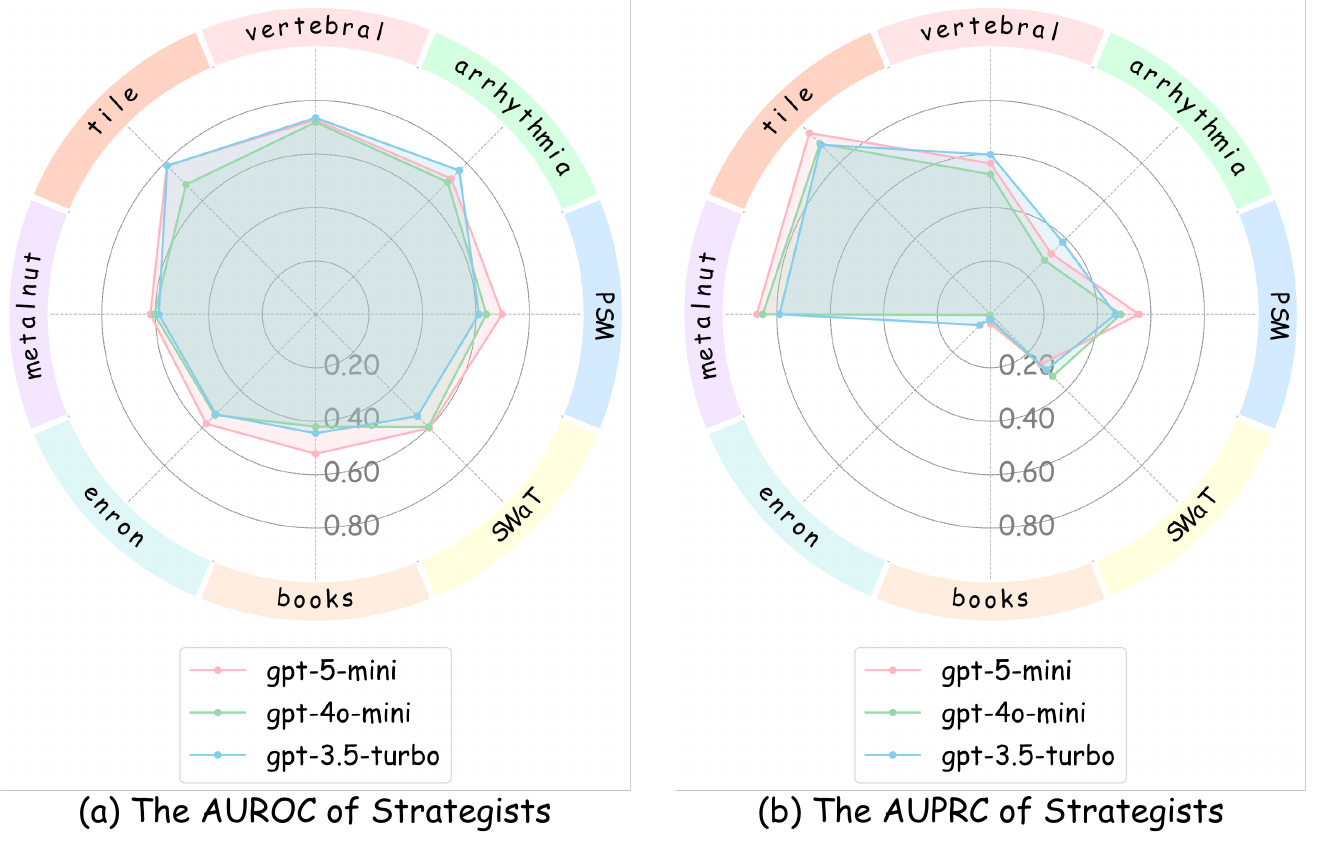}
  \caption{Average Performance of Strategists Utilizing Different LLMs: (a) shows the average values of AUROC on eight datasets and (b) shows the average AUPRC values.}
  \label{fig:llm}
  \vspace{-10pt}
\end{figure}

To evaluate how different language models affect the initial strategy pool, we compare the mean AUROC and AUPRC of strategies generated by the Strategist using gpt 5 mini, gpt 4o mini, and gpt 3.5-turbo across datasets, as shown in Fig.~\ref{fig:llm}.

\textbf{Strategy Quality Depends on Both Model Capability and Data Modality.}
Based on AUROC, gpt 5 mini outperforms gpt 4o mini and gpt 3.5 on most datasets in Fig.~\ref{fig:llm}, including PSM, SWaT, enron, and metalnut. However, this advantage is not universal: on the tabular datasets vertebral and arrhythmia, gpt 3.5 achieves higher average AUROC than gpt 5 mini. Strategy quality also varies significantly across modalities. All models produce strong AUPRC on image datasets, with gpt 5 mini reaching 0.9594 on tile, while the average AUPRC remains very low on the graph datasets, indicating that graph anomaly detection remains challenging. Moreover, AUROC and AUPRC can be inconsistent: gpt 3.5 achieves high AUROC but relatively low AUPRC on arrhythmia, suggesting good ranking ability but weak anomaly precision. 

\section{Conclusion and Future Work}
In conclusion, we propose a DMAIC-inspired multi-agent framework for reliable automated IAD across heterogeneous modalities. 
Following the "Plan First, Judge Later" paradigm, the framework combines SOP distillation with a lightweight execution-free Judge Model for structured planning and pre-execution validation. 
Experiments show that our method achieves state-of-the-art performance and the highest success rate. 
Future work will focus on improving reliability and generalization for safety-critical industrial deployment.

\section{Limitation}
Despite its overall effectiveness, the execution-free Judge Model may still face challenges in out-of-distribution cases within a known modality. This indicates that stronger dataset-aware generalization remains necessary. Besides, this system currently operates as a static framework and does not yet possess continuous self-evolving capabilities. In practical industrial environments, anomaly patterns and data distributions may evolve over time, requiring the system to continuously absorb new feedback, refine its strategy pool, and adapt its decision-making process. Developing such a lifelong self-evolution mechanism constitutes an important direction for future work.

\section{Ethical Considerations}

This work targets agentic industrial anomaly detection, where system outputs may affect quality inspection, maintenance decisions, and operational safety. False negatives may leave faults undetected, while false positives may cause unnecessary production interruptions or maintenance costs. Moreover, LLM-generated strategies and code may still contain incorrect assumptions or implementation errors, particularly under out-of-distribution data. Therefore, generated workflows should be audited and validated by domain experts before deployment. Since industrial data may contain proprietary or sensitive operational information, practitioners should also ensure proper anonymization, access control, and data governance when applying LLM-based agents or external retrieval tools. Our experiments are conducted on public benchmark datasets and do not involve personally identifiable information.

In this paper, AI-assisted technologies were used solely to assist with language polishing and grammar checking. We explicitly state that no AI was utilized to generate research ideas, write the core manuscript, generate citations, or create any images and data. The authors rigorously cross-checked all AI-assisted modifications and take responsibility for the scientific integrity of this manuscript, explicitly ensuring the validity of the empirical findings and the precise accuracy of all references.

%\bibliographystyle{ACM-Reference-Format}
\bibliography{ref}

\appendix

\section*{Appendices}
\addcontentsline{toc}{section}{Appendices}
\startcontents[appendix]

% 打印附录小目录
\subsubsection*{Appendices Content}
\printcontents[appendix]{}{1}{} 

\vspace{1em}

\input{latex/Appendix_experience_reuse}

\input{tables/other_work}

\section{Notation}
\label{Notation}
\input{tables/summary_of_notion}

\section{Evaluation Metrics}
\label{auroc}

\subsection{Area Under the Receiver Operating Characteristic Curve (AUROC)}

The Receiver Operating Characteristic (ROC) curve is a graphical representation used to illustrate the diagnostic ability of a binary classifier system as its discrimination threshold is varied. The curve is created by plotting the True Positive Rate (TPR), also known as recall or sensitivity, against the False Positive Rate (FPR) at various threshold settings. The TPR is the ratio of correctly predicted positive observations to all actual positives, while the FPR is the ratio of incorrectly predicted positive observations to all actual negatives. The Area Under the ROC Curve (AUROC) quantifies the overall performance of the classifier. It represents the probability that the classifier will rank a randomly chosen positive instance higher than a randomly chosen negative one. An AUROC of 1.0 indicates a perfect classifier, while an AUROC of 0.5 suggests a classifier with no discriminative ability, equivalent to random guessing. The AUROC is defined by the integral of the ROC curve:

$$ \text{AUROC} = \int_{0}^{1} \text{TPR}(\text{FPR}^{-1}(t)) \,dt $$

Where $ \text{TPR}(\text{FPR}^{-1}(t)) $ represents the ROC curve.

\subsection{Area Under the Precision-Recall Curve (AUPRC)}

The Precision-Recall (PR) curve is another critical evaluation metric for binary classification, which plots precision against recall (or TPR) for different thresholds. Precision is the ratio of correctly predicted positive observations to the total predicted positives, and recall is the same as TPR. The Area Under the PR Curve (AUPRC) summarizes this plot into a single value. Unlike AUROC, AUPRC is particularly informative when dealing with imbalanced datasets, where the number of negative samples significantly outweighs the number of positive samples. This is because the calculation of precision does not take into account the number of true negatives, focusing instead on the model's performance on the positive class. A higher AUPRC value indicates a better-performing classifier, with 1.0 being the ideal score. The AUPRC is calculated as the area under the PR curve:

$$ \text{AUPRC} = \int_{0}^{1} P(r) \,dr $$

Where $ P(r) $ is the precision as a function of recall $ r $.

\section{Experiment: Comparison with Modality-Specific Baselines}
\label{app:modality_specific_comparison}

To further evaluate the effectiveness of our framework across different data modalities, we compare our method with representative modality-specific anomaly detection baselines. Specifically, we use SVC (One Class SVM) for tabular data, GNN-based models for graph data, LSTM-based models for time-series data, and AutoEncoder-based models for image data. Table~\ref{tab:ours_vs_modality_specific} summarizes the comparison between these modality-specific baselines and our full framework.

As shown in Table~\ref{tab:ours_vs_modality_specific}, our method achieves better performance than modality-specific baselines on most datasets, especially on graph, time-series, and image anomaly detection tasks. Although the SVC baseline slightly outperforms our method on the vertebral dataset, our framework demonstrates more consistent adaptability across heterogeneous modalities without manually selecting a fixed algorithm for each task.

\begin{table*}[t]
\centering
\caption{Comparison with modality-specific baselines across different modalities. All results are reported in percentage (\%). The better result between the modality-specific baseline and our method is highlighted in bold.}
\label{tab:ours_vs_modality_specific}
\resizebox{0.95\linewidth}{!}{
\begin{tabular}{llccccc}
\toprule
\textbf{Modality} & \textbf{Dataset} & \textbf{Baseline} 
& \textbf{Baseline AUROC} & \textbf{Baseline AUPRC} 
& \textbf{Ours AUROC} & \textbf{Ours AUPRC} \\
\midrule
Tabular 
& vertebral  
& SVC         
& \textbf{96.83} & \textbf{91.94} 
& 96.17 & 91.38 \\

Tabular 
& arrhythmia 
& SVC         
& 84.37 & 46.59 
& \textbf{87.99} & \textbf{59.13} \\

Graph 
& books      
& GNN         
& 37.77 & 2.00  
& \textbf{57.55} & \textbf{6.40} \\

Graph 
& enron      
& GNN         
& 76.13 & 0.31  
& \textbf{93.20} & \textbf{1.08} \\

Time Series 
& PSM        
& LSTM        
& 75.87 & 46.88 
& \textbf{87.02} & \textbf{72.60} \\

Time Series 
& SWaT       
& LSTM        
& 50.21 & 5.77  
& \textbf{83.94} & \textbf{67.67} \\

Image 
& metalnut   
& AutoEncoder 
& 54.10 & 68.17 
& \textbf{87.24} & \textbf{96.99} \\

Image 
& tile       
& AutoEncoder 
& 69.34 & 86.76 
& \textbf{70.43} & \textbf{91.81} \\
\bottomrule
\end{tabular}
}
\end{table*}

\section{Experiment: Parameters and Prompt}
\label{parameter experiment}

In order to further analyze the influence of key design choices, we conduct a parameter experiment on the MetalNut dataset. Specifically, we evaluate the effects of prompt design, the number of generated candidate strategies, and the hidden dimension $d_{\text{model}}$ of the Judge Model. The default setting of our framework uses 10 candidate strategies and sets $d_{\text{model}}=128$.

As shown in Table~\ref{tab:parameter_experiment}, the default configuration achieves the best performance, indicating that both sufficient strategy diversity and an appropriately sized Judge Model are important for effective strategy selection. Compared with the simple prompt, the structured prompt used in our framework brings clear improvements, demonstrating the benefit of explicitly guiding the agent with task-specific reasoning procedures. Increasing the number of strategies from 1 to 5 improves performance, while using 10 strategies further enhances the results by providing a richer candidate pool for the Judge Model. For the Judge Model architecture, $d_{\text{model}}=128$ performs better than both smaller and larger dimensions, suggesting that an overly small model may lack sufficient representation capacity, while an overly large model may introduce unnecessary complexity.

\begin{table}[t]
\centering
\caption{Parameter experiment on the MetalNut dataset. We evaluate the effects of prompt design, the number of generated strategies, and the hidden dimension $d_{\text{model}}$ of the Judge Model.}
\label{tab:parameter_experiment}
\resizebox{\linewidth}{!}{
\begin{tabular}{lcc}
\toprule
\textbf{Setting} & \textbf{AUROC (\%)} & \textbf{AUPRC (\%)} \\
\midrule
DMAIC-IAD \\10 Strategies, $d_{\text{model}}=128$ & \textbf{87.24} & \textbf{96.99} \\
Simple Prompt & 80.32 & 87.41 \\
1 Strategy & 69.53 & 83.29 \\
5 Strategies & 74.87 & 87.33 \\
$d_{\text{model}}=64$ & 74.65 & 85.07 \\
$d_{\text{model}}=256$ & 83.11 & 91.26 \\
\bottomrule
\end{tabular}
}
\end{table}

\input{latex/case_study}

\section{Large Language Model Configuration}
\label{sec:llm-config}

The pipeline delegates four language-model stages: pre-analysis, SOP synthesis, meta-planning, and code generation (including per-step agents that share the code-generation backend). Table~\ref{tab:llm-config} summarizes the deployed model identifiers, API vendors, completion budgets, and stage responsibilities. Pre-analysis and SOP generation use \texttt{gpt-4o} via the OpenAI API with a maximum output length of 4{,}096 tokens. The default meta-planner is \texttt{gpt-5-mini} (8{,}192 tokens), which emits structured JSON plans covering preprocessing, feature extraction, the anomaly-detection algorithm, and evaluation metrics. All executable Python---full pipelines and step-level modules---is produced by \texttt{claude-sonnet-4-5} through the Anthropic API (8{,}192 tokens). For each dataset run we generate $K{=}10$ alternative plans; inter-plan diversity is enforced by prompt-level variant constraints and by conditioning each new plan on methods already chosen in previous variants, rather than by varying decoding hyperparameters at the planner.

Sampling controls differ across API generations. Stages implemented with Anthropic Claude and with OpenAI GPT-4o for pre-analysis and SOP use deterministic decoding with temperature $\tau{=}0$. For planner ablations on OpenAI GPT-3.5 and GPT-4--family models (\texttt{gpt-3.5-turbo}, \texttt{gpt-4o-mini}), the provider exposes a temperature parameter; in our experiments we set $\tau{=}1$ (the value supported for these endpoints in our deployment). In contrast, newer OpenAI planner backbones in the GPT-5 family (\texttt{gpt-5-mini}, \texttt{gpt-5.2}) do not accept user-specified temperature in our integration: the client omits the field and the service applies its fixed default decoding policy. We therefore mark planner temperature as not applicable (``--'') in Table~\ref{tab:llm-config} and report planner comparisons primarily in terms of model identity and completion budget (Table~\ref{tab:planner-ablation}), holding non-planner stages fixed.
\input{tables/LLM_parameter}

\input{tables/Appendix_prompts}

\end{document}

%% file: tables/main_experiment.tex
\begin{table*}
\caption{IAD Performance Comparison: We compare our proposed method with three baselines and key ablations. \textit{+ wo Judge Model} means our framework without judge model and \textit{+ wo SOP} means our framework without Consultant Agent. "$\uparrow$" means the larger the better and "/" means this method can not solve the task. The evaluation is conducted on eight benchmark datasets using AUROC and AUPRC as the primary metrics. All the results are the average value of 5 repeated experiments. The best results are highlighted in bold.}
\label{tab:main}
\resizebox{\textwidth}{!}{%
\begin{tabular}{ccccccccc}
\hline
                             & \multicolumn{4}{c}{Tabular Data}                                      & \multicolumn{4}{c}{Time Series Data}                         \\ \cline{2-9} 
                             & \multicolumn{2}{c}{vertebral}     & \multicolumn{2}{c}{arrhythmia}    & \multicolumn{2}{c}{PSM}           & \multicolumn{2}{c}{SWaT} \\ \hline
Method                       & AUROC$\uparrow$          & AUPRC$\uparrow$           & AUROC$\uparrow$          & AUPRC$\uparrow$           & AUROC$\uparrow$          & AUPRC$\uparrow$           & AUROC$\uparrow$      & AUPRC$\uparrow$      \\ \hline
AD-AGENT                     & 0.7024          & 0.3273          & 0.6750          & \textbf{0.6030}          & 0.5217          & 0.3981          & 0.5797      & 0.0802     \\
AutoIAD                      & /               & /               & /               & /               & /               & /               & /           & /          \\
Strategist   only            & 0.6190          & 0.5601          & 0.7772          & 0.2647          & 0.6888          & 0.3744          & 0.3839      & 0.0435     \\
\textbf{DMAIC-IAD}                         & \textbf{0.9617} & \textbf{0.9138} & \textbf{0.8799} & 0.5913 & \textbf{0.8702}         & \textbf{0.7260}          & \textbf{0.8394}      & \textbf{0.6767}     \\
\textit{+ wo Judge Model} & 0.7276           & 0.5650           & 0.7181          & 0.3211          & 0.6981          & 0.5555          & 0.6023      & 0.2651     \\
\textit{+ wo SOP}         & 0.7962          & 0.6936          & 0.7357          & 0.4239          & 0.5000          & 0.3063          & 0.5000      & 0.0581     \\ \hline
                             & \multicolumn{4}{c}{Graph   Data}                                      & \multicolumn{4}{c}{Image Data}                               \\ \cline{2-9} 
                             & \multicolumn{2}{c}{books}         & \multicolumn{2}{c}{enron}         & \multicolumn{2}{c}{metalnut}      & \multicolumn{2}{c}{tile} \\ \hline
Method                       & AUROC$\uparrow$          & AUPRC$\uparrow$           & AUROC$\uparrow$          & AUPRC $\uparrow$          & AUROC$\uparrow$          & AUPRC$\uparrow$          & AUROC$\uparrow$      & AUPRC$\uparrow$      \\ \hline
AD-AGENT                      & 0.5235          & 0.0265          & 0.5000          & 0.002          & /               & /               & /           & /          \\
AutoIAD                      & /               & /               & /               & /               & 0.8000          & 0.8583          & 0.7338      & 0.8944     \\
Strategist only              & 0.4616          & 0.0355          & 0.1160          & 0.0007          & 0.6466 &	0.9058             & 0.5783      & 0.7606     \\
\textbf{DMAIC-IAD}                        & \textbf{0.5755} & \textbf{0.0640} & \textbf{0.9320} & \textbf{0.0108} & \textbf{0.8724} & \textbf{0.9699} & 0.7043      & 0.9181     \\
\textit{+ wo Judge Model} & 0.5220          & 0.0363          & 0.5790          & 0.0044          & 0.7043          & 0.9181          & \textbf{0.7873}      & \textbf{0.9594}     \\
\textit{+ wo SOP}         & 0.4964          & 0.0268          & 0.4425          & 0.0017          & 0.5000          & 0.8087          & 0.4098      & 0.6765     \\ \hline
\end{tabular}
}
\vspace{-10pt}
\end{table*}

%% file: tables/success_rate_tabel.tex
% 正文中：
\begin{table}
\centering
\caption{Comparison of Task Completion and Cost: This table assesses the success rate (SR) (\%), execution time, Completion Token (CT) and Prompt Token (PT) of our proposed method against three baselines and two ablation variants, \textit{+ wo Judge Model} and \textit{+ wo SOP}. $\uparrow$ means the higher the better and $\downarrow$ means the lower the better. The highest success rate is highlighted.}
\label{tab:success_rate}
\resizebox{\columnwidth}{!}{%
\begin{tabular}{@{}ccccc@{}}
\toprule
Model   Name                 & SR$\uparrow$     & Time(s)$\downarrow$ & CT$\downarrow$ & PT$\downarrow$ \\ \midrule
AD-AGENT                      & 42.76\%          & 57.75   & 7,892             & 5,863         \\
AutoIAD                      & 74.36\%          & 108.43  & 96,906            & 65,348        \\
Strategist only              & 53.90\%          & 125.07  & 13,446             & 8,732        \\
\textbf{DMAIC-IAD}                         & \textbf{78.19\%} & 298.37  & 20,467            & 15,320        \\
\textit{+ wo Judge Model} & 77.23\%          & 153.77  & 14,868            & 10,358        \\
\textit{+ wo SOP}         & 69.33\%          & 219.62  & 16,867            & 12,499        \\ \bottomrule
\end{tabular}
}
\vspace{-10pt}
\end{table}

%% file: latex/Appendix_experience_reuse.tex
\section{Implementation Details of Efficient Experience Reuse}
\label{app:experience_reuse}

This section further clarifies the scope and implementation of the experience reuse mechanism introduced in Section~\ref{define}. The memory module is implemented as an exact-match hash table rather than a vector database. Given a normalized task description $r$ and dataset schema $D$, the system computes deterministic fingerprints $H(r)$ and $H(D)$ and uses them as lookup keys for previously generated SOPs and data profiles, respectively.

The reuse mechanism follows a lookup-first protocol. If $H(r)$ yields a cache hit, the corresponding SOP $w$ is directly loaded, and the Consultant Agent is skipped. If $H(D)$ yields a cache hit, the stored data profile $\mathbf{m}$ is reused, and the Data Inspector Agent is skipped. If either key is not found, the corresponding artifact is regenerated through the standard DMAIC workflow and then inserted into memory for future reuse. This design also supports partial reuse: for example, an existing SOP can be reused for a recurring industrial scenario even when the dataset schema is new, while a data profile can be reused when the same dataset schema appears under a slightly different task description.

The claimed $O(1)$ complexity specifically refers to the average-case retrieval complexity of the proposed hash-based experience reuse mechanism, rather than the runtime complexity of the entire pipeline. In our lightweight mode, each recurring task is represented by deterministic fingerprints of the task description and dataset schema, denoted as $H(r)$ and $H(D)$, respectively. These fingerprints are used as keys in a hash-table memory:
\[
\mathcal{M} = \{H(r) \mapsto w\} \cup \{H(D) \mapsto \mathbf{m}\},
\]
where $w$ denotes the previously generated SOP and $\mathbf{m}$ denotes the corresponding data profile. Given a new task, our system first computes its fingerprints and directly queries the hash table. Under the standard average-case assumption of hash-table access, checking whether $H(r)$ or $H(D)$ exists in memory and retrieving the associated SOP or data profile requires constant time, i.e., $O(1)$, independent of the number of previously stored experiences.

%% file: tables/other_work.tex
\section{Related Work}
\label{appendix:related work}
\begin{table*}[ht]
\centering
\caption{Comparison of anomaly detection agents across modalities and modulars.
\cmark: explicitly supported and evaluated; 
\xmark: not supported or not considered.}
\label{tab:modality_benchmark_comparison}
\resizebox{2\columnwidth}{!}{
\begin{tabular}{@{}ccccc@{}}
\toprule
                                                   &                     & AD-AGENT \cite{yang2025ad} & AutoIAD \cite{ji2025autoiad} & Ours \\ \midrule
\rowcolor[HTML]{ECF4FF} 
\cellcolor[HTML]{ECF4FF}                           & Time Series         & \cmark        & \xmark       & \cmark    \\
\rowcolor[HTML]{ECF4FF} 
\cellcolor[HTML]{ECF4FF}                           & Tabular             & \cmark        & \xmark       & \cmark    \\
\rowcolor[HTML]{ECF4FF} 
\cellcolor[HTML]{ECF4FF}                           & Graph               & \cmark        & \xmark       & \cmark    \\
\rowcolor[HTML]{ECF4FF} 
\multirow{-4}{*}{\cellcolor[HTML]{ECF4FF}Modality} & Image               & \xmark        & \cmark       & \cmark    \\ \midrule
\rowcolor[HTML]{FFFFE1} 
\cellcolor[HTML]{FFFFE1}                           & Knowledge           & \xmark        & \xmark       & \cmark    \\
\rowcolor[HTML]{FFFFE1} 
\cellcolor[HTML]{FFFFE1}                           & Data Profile        & \xmark        & \cmark       & \cmark    \\
\rowcolor[HTML]{FFFFE1} 
\cellcolor[HTML]{FFFFE1}                           & Strategy Generation & \cmark        & \cmark       & \cmark    \\
\rowcolor[HTML]{FFFFE1} 
\cellcolor[HTML]{FFFFE1}                           & Strategy Evaluation & \xmark        & \xmark       & \cmark    \\
\rowcolor[HTML]{FFFFE1} 
\multirow{-5}{*}{\cellcolor[HTML]{FFFFE1}Modular}  & Code Generator      & \cmark        & \cmark       & \cmark    \\ \bottomrule
\end{tabular}}
\end{table*}

\subsection{Conventional Industrial Anomaly Detection}
%缩短，原版放到table.tex的最后面了
%加了一个总结句，把原本的两段Reconstruction-based methods和Feature embedding-based methods合并了
Industrial anomaly detection has been extensively studied, particularly under unsupervised settings where only normal data are available for training \cite{chalapathy2019deep, ruff2021unifying}.Existing methods can be broadly categorized into reconstruction-based and feature embedding-based approaches. Reconstruction-based methods detect anomalies through reconstruction or prediction errors using autoencoders, recurrent models, or, more recently, diffusion models \cite{sakurada2014anomaly, park2018multimodal, wyatt2022anoddpm}. Feature embedding-based methods identify anomalies by measuring deviations in learned representation spaces, such as memory bank approaches and normalizing flow-based density estimation \cite{roth2022towards, kirichenko2020normalizing}. Beyond these paradigms, forecasting-based and graph-based models have also been explored to capture temporal dynamics and relational dependencies in industrial systems \cite{hundman2018detecting, lv2023adaptive}. 

%other approaches和%summary也合并了
Despite strong performance in constrained environments, most deep learning-based industrial anomaly detection methods remain model-centric and scenario-dependent, often requiring careful model design and retraining for each specific setting. This limits their scalability, adaptability, and generalization across heterogeneous industrial scenarios \cite{chalapathy2019deep, li2025survey}.

\subsection{Agentic System for Anomaly Detection}
Early agent-based methods for anomaly detection focused on distributed monitoring, where agents observe system states and trigger alarms upon detecting abnormal behaviors  \cite{mcarthur2005agent}. Subsequent works extended this paradigm to multi-agent systems, enabling collaboration among specialized agents to improve robustness and scalability in complex industrial environments \cite{garcia2019multi}.

More recently, large model empowered agentic systems have emerged as a promising paradigm, enabling agents to reason, plan, and collaborate \cite{su2025many}. There are some works that leverage LLMs or LVMs for IAD. In the time series area, Argos \cite{gu2025argos} utilizes LLMs for autonomous rule generation, ADT \cite{yang2025agent} uses reinforcement learning to realize thresholding control, and MAS-LSTM \cite{qin2025mas} proposes a LSTM-based multi-agent system for IoT datasets. As for the vision scenario, AgentIAD \cite{miao2025agentiad} uses LVM for multi-stage visual inspection. Besides, LEMAD \cite{ji2025lemad} proposes a multi-agent system for Power Grid Services to process log anomaly detection. PARAM \cite{harbola2025prescriptive}uses rag for data knowledge retrieval and utilizes a pretrained LLM to realize anomaly detection. These works use LLM as a specific processor without considering using a specific model for different scenarios.

Representative frameworks such as AD-AGENT \cite{yang2025ad} and AutoIAD \cite{ji2025autoiad} construct end-to-end pipelines that autonomously generate executable code for diverse industrial scenarios. As shown in Table. \ref{tab:modality_benchmark_comparison}, AD-AGENT proposes a selection model, then code generation and optimizing parameters. It focuses on the optimization step to get better parameters. AutoIAD decomposes the data analysis process into its work, augmenting data preparation, data loading, model design, and the training process. They put their attention on the realization process, instead of decision-making.

%% file: tables/summary_of_notion.tex
\begin{table}[htbp]
\centering
\caption{Summary of Notation in the DMAIC-Agentic Framework}
\label{tab:notations}
\renewcommand{\arraystretch}{1.2}
\resizebox{\linewidth}{!}{
\begin{tabular}{l|l}
\hline
\textbf{Notation} & \textbf{Description} \\
\hline
\multicolumn{2}{c}{\textit{Sets and Domains}} \\
\hline
$\Omega$ & The global tuple defining the IAD Agentic Framework \\
$\mathcal{R}$ & Domain of task descriptions (User Scenarios) \\
$\mathcal{D}$ & Domain of datasets \\
$\mathcal{K}$ & External knowledge base (Literature \& Web Resources) \\
$\mathcal{P}$ & The ordered set of DMAIC phases \\
$\mathcal{W}$ & Domain of Standard Operating Procedures (SOPs) \\
$\mathcal{M}$ & Space of structured data profiles \\
$\Pi$ & Set of executable strategies \\
$\Pi^K$ & Set of K candidate strategies generated by the Strategist \\
$\mathcal{C}$ & Space of executable code scripts \\
\hline
\multicolumn{2}{c}{\textit{Functions and Operators}} \\
\hline
$\Psi_{cons}$ & Consultant function: Retrieval and SOP distillation \\
$\Psi_{insp}$ & Inspector function: Feature extraction and profiling \\
$\Psi_{strat}$ & Strategist function: Candidate strategy generation \\
$\Psi_{code}$ & Executor function: Code generation from strategy \\
$\Psi_{refine}$ & Code refinement function based on feedback \\
$\Phi_{judge}$ & Judge utility function for scoring strategies \\
$\mathcal{F}_{mlp}$ & MLP regressor function within the Judge Model \\
$V(\cdot)$ & Validation oracle for code execution \\
$E(\cdot)$ & Semantic embedding function (Sentence Transformer) \\
$H(\cdot)$ & Deterministic hash function for fingerprinting \\
$\oplus$ & Vector concatenation operator \\
\hline
\multicolumn{2}{c}{\textit{Variables and Artifacts}} \\
\hline
$r$ & Specific input task description ($r \in \mathcal{R}$) \\
$w$ & Structured SOP, composed of $\{d_{scen}, \vec{o}_{steps}, \mathcal{M}_{rec}\}$ \\
$D$ & Raw input dataset \\
$\mathbf{m}$ & Structured data profile, composed of $\langle \phi_{feat}, \tau_{task}, \delta_{miss} \rangle$ \\
$K$ & The number of candidate strategies \\
$L_k$ & The length (number of steps) of strategy $\pi_k$ \\
$\pi_k$ & The $k$-th candidate strategy ($\pi_k \in \Pi$) \\
$\tau_j$ & A single step tuple $\langle i, \alpha_{act}, \mu_{algo}, \theta_{param} \rangle$ in a strategy \\
$\pi^*$ & The optimal strategy selected by the Judge \\
$\hat{y}_k$ & Predicted utility score for strategy $\pi_k$ \\
$\theta$ & Trainable parameters of the Judge Model's MLP \\
$c^{(t)}$ & Executable code script at iteration $t$ \\
$\epsilon^{(t)}$ & Error trace or feedback from validation \\
$T_{max}$ & Maximum number of iterations for code refinement \\
$\mu^{(hist)}$ & Historical performance metrics (Ground Truth) \\
$\lambda_j$ & Importance weight for the $j$-th historical metric \\
\hline
\end{tabular}
}
\end{table}

%% file: latex/case_study.tex
\section{Case Study}
\label{case study}

% In your document body, where you want the example to appear.
% Using figure* makes the box span both columns, which is recommended for readability.
% If you want it strictly in one column, use \begin{figure}[h!] and \end{get}{figure}.
% In your document body, where you want the example to appear.
% Using figure* makes the box span both columns, which is recommended for readability.

\begin{figure*}[t]
  \vspace{-20pt}
\begin{tcolorbox}[
    colback=orange!5!white, % Light orange background
    colframe=orange!80!black, % Dark orange frame
    title=Literature-Based SOP for MVTec Tile Dataset,
    fonttitle=\bfseries,
    left=6mm,
    enhanced,
    attach boxed title to top left={xshift=6mm, yshift=-2mm},
    boxed title style={
        colback=orange!80!black,
        sharp corners,
        top=1mm,
        bottom=1mm,
    }
]
\begin{small} % Use a smaller font size for the content

% User Requirement Summary
\textbf{Requirement Summary:} Analyze the MVTec Tile dataset for texture-based anomaly detection.
\begin{itemize}[leftmargin=*,topsep=2pt,itemsep=1pt]
    \item \textbf{Type:} Building materials / Surface texture.
    \item \textbf{Anomalies:} Cracks, bubbles or pits, color differences, edge chipping.
    \item \textbf{Characteristics:} Normal samples have repetitive patterns; anomalies are structural damage. Suited for feature matching or reconstruction methods.
\end{itemize}

\hrulefill % Separator line

% Generated SOP
\textbf{Generated Standard Operating Procedure (SOP):}

\begin{itemize}[leftmargin=*,topsep=3pt,itemsep=2pt]
    \item \textbf{Scenario:} Anomaly detection for tile surface texture dataset.
    \item \textbf{Data Description:} Image dataset of tile surfaces with repetitive patterns. Normal samples exhibit consistent textures, while anomalies include structural damages.
\end{itemize}

\textbf{Operation Steps:}
\begin{enumerate}[leftmargin=*,label=\textbf{Step \arabic*:},wide, topsep=4pt,itemsep=4pt]
    \item \textbf{Data Preprocessing and Tiling:} Segment the input images into smaller tiles to enhance the detection of local anomalies, as described in.
    \item \textbf{Texture Feature Extraction:} Extract texture features using procedural texturing workflows and stochastic texturing methods to capture repetitive patterns.
    \item \textbf{Reconstruction-Based Anomaly Detection:} Implement reconstruction-based methods, training a model on normal samples. Techniques like tiled diffusion can be adapted.
    \item \textbf{Feature Matching for Anomaly Detection:} Apply feature matching to compare test sample features against a memory bank of normal features, potentially using tools like Bitmap2Material.
    \item \textbf{Calibration and Evaluation:} Calibrate the detection system, drawing inspiration from methods like the ATLAS Tile Calorimeter calibration [Paper Search Results], and evaluate using statistical metrics.
\end{enumerate}

\textbf{Recommendations:}
\begin{itemize}[leftmargin=*,topsep=3pt,itemsep=2pt]
    \item \textbf{Preprocessing:} Use tiling to segment images for texture-based models.
    \item \textbf{Feature Extraction:} Leverage procedural and stochastic texturing methods.
    \item \textbf{Modeling:} Adopt a hybrid approach combining reconstruction-based (e.g., tiled diffusion) and feature-matching methods.
    \item \textbf{Evaluation:} Calibrate the system and evaluate using reconstruction error and feature matching accuracy.
\end{itemize}
\end{small}
\end{tcolorbox}
\vspace{-10pt}
\caption{An example of a SOP generated for the Tile dataset.}
\label{fig:sop_tile}
  \vspace{-10pt}
\end{figure*}

% In your document body, where you want the example to appear.
% Using figure* makes the box span both columns, which is recommended for readability.
\begin{figure*}[!]
  
\begin{tcolorbox}[
    colback=purple!5!white, % Light purple background
    colframe=purple!75!black, % Dark purple frame
    title=Data Profile for MVTec Tile Dataset,
    fonttitle=\bfseries,
    left=6mm,
    enhanced,
    attach boxed title to top left={xshift=6mm, yshift=-2mm},
    boxed title style={
        colback=purple!75!black,
        sharp corners,
        top=1mm,
        bottom=1mm,
    }
]
\begin{small} % Use a smaller font size for the content

% User Requirement Summary
\textbf{Requirement Summary:} Analyze the MVTec Tile dataset for texture-based anomaly detection.
\begin{itemize}[leftmargin=*,topsep=2pt,itemsep=1pt]
    \item \textbf{Type:} Building materials / Surface texture.
    \item \textbf{Anomalies:} Cracks, bubbles or pits, color differences, edge chipping.
    \item \textbf{Characteristics:} Normal samples exhibit repetitive patterns, while anomalies are mainly structural damage. Suited for feature matching or reconstruction methods.
\end{itemize}

\hrulefill % Separator line

% Generated Data Profile
\textbf{Generated Data Profile:}

\begin{itemize}[leftmargin=*,topsep=3pt,itemsep=2pt]
    \item \textbf{Task Type:} \texttt{classification} (Binary)
    \item \textbf{Data Type:} \texttt{Image}
    \item \textbf{Supervision:} Supervised (due to presence of labels).
    \item \textbf{Label Information:}
        \begin{itemize}[leftmargin=10mm,label=\textendash,topsep=1pt,itemsep=1pt]
            \item \textbf{Training Set:} Contains 230 samples, all are normal (label 0).
            \item \textbf{Test Set:} Contains 117 samples, composed of 33 normal (label 0) and 84 anomalous (label 1) samples.
        \end{itemize}
    \item \textbf{Keywords:} \texttt{tile}, \texttt{surface}, \texttt{anomaly}, \texttt{texture}, \texttt{detection}
    \item \textbf{Insights Summary:}
    \begin{quote}
        \textit{The dataset is structured for a binary classification task where the training set contains only normal samples, and the test set is imbalanced with both normal and anomalous samples. This setup is typical for evaluating supervised anomaly detection methods that learn a model of normalcy from a "clean" training distribution. Although the domain is texture anomaly detection, the availability of explicit labels defines the technical task as classification.}
    \end{quote}
\end{itemize}
\end{small}
\end{tcolorbox}
\caption{The data profile automatically generated by our Data Inspector agent for the MVTec Tile dataset. It correctly identifies the task structure, label distribution, and key characteristics, providing a foundation for subsequent strategy generation.}
\label{fig:data_profile_tile}
\end{figure*}

\begin{figure*}[t!]
\begin{tcolorbox}[
    colback=green!5!white, % Light green background
    colframe=green!65!black, % Dark green frame
    title=Example of an LLM-Generated Execution Strategy (for Tile Dataset),
    fonttitle=\bfseries,
    left=6mm,
    enhanced,
    attach boxed title to top left={xshift=6mm, yshift=-2mm},
    boxed title style={
        colback=green!65!black,
        sharp corners,
        top=1mm,
        bottom=1mm,
    }
]
\begin{small} % Use a smaller font size for the content

% User Requirement Section
\textbf{User Requirement:} Analyze the Tile image anomaly detection dataset from MVTec AD.
\begin{itemize}[leftmargin=*,topsep=2pt,itemsep=1pt]
    \item \textbf{Dataset Type:} Texture category (building materials).
    \item \textbf{Image Resolution:} Approx. 640$\times$640 pixels.
    \item \textbf{Common Anomaly Types:} Cracks, bubbles or pits, color differences, edge chipping.
    \item \textbf{Characteristics:} Normal samples have repetitive patterns; anomalies are mostly structural. Suitable for reconstruction-based or feature matching methods.
\end{itemize}

\hrulefill % Separator line

% Steps Section
\textbf{Generated Execution Steps:}

\begin{enumerate}[leftmargin=*,label=\textbf{Step \arabic*:},wide, topsep=4pt,itemsep=4pt]

    % STEP 1
    \item \textbf{Data Preprocessing}
    \begin{itemize}[leftmargin=10mm,label=\textbullet,topsep=2pt,itemsep=1pt]
        \item \textbf{Method:} \texttt{Tiling\_128x128\_stride64}
        \item \textbf{Parameters:}
        \begin{itemize}[leftmargin=5mm,label=\textendash,topsep=1pt,itemsep=1pt]
            \item Tile Size: [128, 128], Stride: [64, 64] (50\% overlap).
            \item \textit{Notes:} Segment image into overlapping tiles to capture anomalies spanning boundaries. Tile coordinates are kept for later score aggregation.
        \end{itemize}
    \end{itemize}

    % STEP 2
    \item \textbf{Feature Extraction}
    \begin{itemize}[leftmargin=10mm,label=\textbullet,topsep=2pt,itemsep=1pt]
        \item \textbf{Method:} \texttt{WideResNet50\_patch\_features}
        \item \textbf{Parameters:}
        \begin{itemize}[leftmargin=5mm,label=\textendash,topsep=1pt,itemsep=1pt]
            \item Backbone: WideResNet50 (pretrained on ImageNet).
            \item Output Layer: layer3 (1024-dim).
            \item Pooling: Spatial-average pooling.
            \item \textit{Notes:} Extract per-tile deep patch descriptors to capture local texture patterns, leveraging pretrained feature priors.
        \end{itemize}
    \end{itemize}

    % STEP 3
    \item \textbf{Anomaly Algorithm}
    \begin{itemize}[leftmargin=10mm,label=\textbullet,topsep=2pt,itemsep=1pt]
        \item \textbf{Method:} \texttt{PatchCore}
        \item \textbf{Parameters:}
        \begin{itemize}[leftmargin=5mm,label=\textendash,topsep=1pt,itemsep=1pt]
            \item Memory Bank: Built via coreset sampling (1\% ratio) from normal tiles.
            \item Distance Metric: Cosine distance with k=1 neighbor.
            \item Thresholding: Calibrated on a validation set of normal tiles (99th percentile).
            \item Aggregation: Per-tile scores aggregated to image-level via max-pooling.
            \item \textit{Notes:} Use Faiss for fast nearest neighbor search. Train the memory bank on features from normal tiles only.
        \end{itemize}
    \end{itemize}

\end{enumerate}
\end{small}
\end{tcolorbox}
\caption{A detailed execution strategy generated by our framework for the MVTec Tile dataset. The plan specifies a concrete sequence of methods, from tiling and feature extraction with WideResNet50 to anomaly detection using PatchCore, including key hyperparameters.}
\label{fig:plan_example_tile}
\end{figure*}
In the Define phase, the IAD Consultant identifies the MVTec Tile dataset as a texture-based surface inspection task by retrieving domain-specific knowledge. As illustrated in the SOP Display (Fig. \ref{fig:sop_tile}), the generated Standard Operating Procedure (SOP) formalizes essential workflows tailored to tile surfaces, prioritizing structural damage detection (e.g., cracks, pits) through a combination of procedural tiling and hybrid reconstruction-feature matching strategies.

During the Measure phase, the Data Inspector profiles the dataset, confirming its status as an unsupervised learning problem characterized by a "clean" training set composed entirely of normal samples (as detailed in Fig. \ref{fig:data_profile_tile}). This profiling clarifies the task structure, confirming that despite the presence of labels, the dataset is optimized for learning normalcy from a distribution of anomaly-free samples.

Based on these constraints, the Analyze phase involves the Strategist formulating a concrete execution plan. As shown in Fig. \ref{fig:plan_example_tile}, the strategy leverages a PatchCore framework, specifically configured with a WideResNet50 backbone to balance spatial resolution and high-level semantics. Key hyperparameters, such as $128 \times 128$ overlapping tiling and the utilization of Layer 3 features, are explicitly defined to optimize local texture analysis.

The Improve phase then features an execution-free Judge Model that evaluates these candidate plans, validating the PatchCore framework for its computational efficiency and suitability for this specific texture profile without requiring costly runtime trials. Finally, the Control phase concludes with the IAD Executor generating the corresponding executable code and a comprehensive, auditable detection report.

%% file: tables/LLM_parameter.tex
\begin{table*}[htbp]
  \centering
  \caption{Large language model settings for the multi-agent anomaly-detection pipeline.}
  \label{tab:llm-config}
  \resizebox{\textwidth}{!}{
  \begin{tabular}{@{}l l l r c p{0.22\linewidth}@{}}
    \toprule
    \textbf{Pipeline stage} & \textbf{Model ID} & \textbf{API} & \textbf{Max.\ output tokens} & \textbf{Temp.\ ($\tau$)} & \textbf{Role} \\
    \midrule
    Data Inspector
      & \texttt{gpt-4o}
      & OpenAI
      & 4{,}096
      & 0.0
      & Infer task type, schema, and data-quality summary from user requirement and sample statistics. \\
    IAD Consultant
      & \texttt{gpt-4o}
      & OpenAI
      & 4{,}096
      & 0.0
      & Produce method-level SOP guidance conditioned on pre-analysis. \\
    Strategist (default)
      & \texttt{gpt-5-mini}
      & OpenAI
      & 8{,}192
      & -
      & Emit JSON execution plans (preprocessing, features, algorithm, metrics). \\
    Code generation agent
      & \texttt{claude-sonnet-4-5}
      & Anthropic API
      & 8{,}192
      & 0.0
      & Generate standalone Python implementing each plan; step agents share this backend. \\
    \midrule
    \multicolumn{6}{@{}l}{\textbf{Multi-plan setting:} $K=10$ plans per run.} \\
    \bottomrule
  \end{tabular}
  }
\end{table*}

\begin{table}[t]
  \centering
  \caption{Planner-only LLM variants for ablation (other stages fixed as in Table~\ref{tab:llm-config}).}
  \label{tab:planner-ablation}
  \begin{tabular}{@{} l r @{}}
    \toprule
    \textbf{Deployed model ID} & \textbf{Max.\ output tokens} \\
    \midrule
    \texttt{gpt-4o-mini}  & 8{,}192 \\
    \texttt{gpt-5-mini}   & 8{,}192 \\
    \texttt{gpt-5.2}      & 8{,}192 \\
    \texttt{gpt-3.5-turbo}& 8{,}192 \\
    \bottomrule
  \end{tabular}
\end{table}

%% file: tables/Appendix_prompts.tex
\section{Agent Prompt}

\subsection{IAD Data Inspector}
\begin{lstlisting}[style=promptstyle]
System Prompt:
You are a data analysis expert. Infer task_type from user requirement + data samples.

RULE: Data structure first. Task type by label structure:
- train_label_unique=1 class AND test_label_unique=2 classes -> anomaly_detection (one-class: train only normal, test has normal+anomaly)
- One label column, >2 classes -> multi_class_classification
- Multiple binary label columns -> multi_label_classification
- One binary label column (both train & test have 2 classes) -> classification
- Continuous target -> regression
- No label column / unlabeled -> anomaly_detection

Output JSON only: {"task_type": "...", "scenario_keywords": ["kw1", ...], "additional_info": "..."}
scenario_keywords: 3-5 domain keywords for paper search. No extra keys, no text outside JSON.
\end{lstlisting}

\begin{lstlisting}[style=promptstyle]
[System]
{self._system_prompt}

[User]
User requirement description:
{requirement}

Data information:
{data_info if data_info else "No data provided"}

Important: Please determine the task type based on the DATA STRUCTURE first (especially whether there are label columns and how many distinct label values), then consider the requirement description.

Output JSON: task_type, is_supervised, scenario_keywords (3-5 domain keywords), additional_info. Output only JSON.
\end{lstlisting}

\subsection{IAD Consultant Agent Prompt}

\begin{lstlisting}[style=promptstyle]
You are an SOP (Standard Operating Procedure) generator for an industrial anomaly detection system.
Task: Generate standardized operating guidelines (SOP guide) based on search results (web search and paper search) and input information.
Output format: JSON format containing structured operation steps and recommendations.

CRITICAL REQUIREMENTS:
- You MUST base the SOP guide PRIMARILY on the provided search results (web search and paper search results)
- The operation steps should reflect methods, workflows, and approaches found in the search results
- Steps can be arranged flexibly based on the search results - you do NOT need to follow a fixed order (preprocessing -> feature extraction -> anomaly detection)
- You can include specific methods or techniques mentioned in the search results
- DO NOT output generic or irrelevant information like 'precautions' or 'best_practices' that are not directly related to the search results
- Focus on actionable, specific steps that reflect the knowledge from search results

Output JSON structure:
{
  "scenario": "brief scenario description",
  "data_description": "brief data description",
  "operation_steps": [
    {
      "step_id": "step_1",
      "step_name": "step name (should reflect methods from search results)",
      "description": "detailed description based on search results",
      "order": 1
    }
  ],
  "recommendations": {
    "data_preprocessing": "recommendations based on search results (if applicable)",
    "feature_extraction": "recommendations based on search results (if applicable)",
    "modeling": "recommendations based on search results (if applicable)",
    "evaluation": "recommendations based on search results (if applicable)"
  }
}

IMPORTANT: Do NOT include 'precautions' or 'best_practices' fields in the output. Only output the fields specified above.

CONSTRAINTS for one-class / MVTec-style image anomaly:
- scenario: Use 'One-class anomaly detection' or 'Unsupervised image anomaly detection', NOT 'Classification anomaly detection'
- data_description: If train has 1 class and test has 2 classes, state: train=normal only, test=normal+anomaly
- feature_extraction: Prefer PatchCore-style, pretrained CNN patch features. Do NOT recommend 'multi-scale self-referencing templates'
- modeling: Prefer PatchCore, kNN on patch features, OneClassSVM. Do NOT recommend YOLOv5, YOLOv8 or object detection models
- When train 1 class and test 2 classes: force anomaly_detection task, avoid classification/detection models
\end{lstlisting}

\subsection{IAD Strategist Prompt}
\begin{lstlisting}[style=promptstyle]
The lists below are suggestions/references. You are free to use any method that fits the task and data (from any library: sklearn, pyod, xgboost, tslib, darts, etc.). Some plans may already follow SOP; you can also choose other methods as you see fit. The only hard rule: for graph data do NOT use PyGOD (pygod); use graph_statistical_features + PyOD/sklearn or alternatives.

Reference: Feature Extraction Methods (by data type)
- Image datasets (MVTec AD, hazelnut, screw, metal_nut, tile, transistor, etc.): PatchCore, CNN_backbone, pretrained CNN patch features. Do NOT use YOLOv5/YOLOv8 or object detection models. MVTec AD is one-class: train=normal only, test=normal+anomaly.
- Tabular/high-dim: PCA, statistical_features: use when dimensionality reduction or statistical aggregation is beneficial.
- Graph data: graph_statistical_features: extract node/graph-level features (degree, clustering coefficient, centrality via NetworkX) and treat as tabular; then use any anomaly detection library (PyOD, sklearn, etc.). DO NOT use PyGOD (pygod) package.
- Time series: PREFER "none" for feature_extraction. TSLib models (TimesNet, FEDformer, Autoformer, etc.) have built-in feature learning and do NOT require external feature extraction or window-based statistical features. Only use time_series_features/window-based features if NOT using TSLib models.
- Generic: none: pass through raw features when no special extraction is needed.

Reference: Methods by Task Type
For Anomaly Detection Tasks (task_type: anomaly_detection)
- Image (MVTec AD style): PatchCore, kNN on patch features, OneClassSVM. Do NOT use YOLOv5/YOLOv8 or object detection.
1. PyOD (Library: pyod)
Applicable Data Type: Multivariate Tabular Data, Graph Data (after feature extraction)
Description: Python Outlier Detection: a comprehensive library with 40+ anomaly detection algorithms
For tabular/numeric data prefer classic, well-established methods: IsolationForest, LOF, OneClassSVM, COPOD, ECOD, ABOD, CBLOF, HBOS (fast and robust). Deep methods (AutoEncoder, VAE, DeepSVDD) are available for special cases where classic methods are insufficient.
Available Models: IsolationForest, LOF, OneClassSVM, COPOD, ECOD, ABOD, AutoEncoder, VAE, DeepSVDD, IForest, CBLOF, HBOS, FeatureBagging, LSCP, SUOD
For Graph Data: Use graph_statistical_features (NetworkX: degree, clustering, PageRank, etc.) to convert graph to tabular, then apply PyOD, sklearn, or other anomaly detection models. AVOID PyGOD (pygod): do not use pygod package.
2. TSLib (Library: tslib) - PREFERRED FOR TIME SERIES DATA
Applicable Data Type: Time Series Data (msl, yahoo, psm, smap, smd, swat, etc.)
Description: Deep learning library for time series developed by THUML. These models have built-in feature learning and do NOT require preprocessing or feature extraction steps.
PRIORITY MODELS (use these first):
- TimesNet - Default choice for general-purpose time series modeling, strong generalization across all tasks
- FEDformer - Frequency-domain decomposition with Transformer, captures global seasonal patterns efficiently
- Autoformer - Auto-correlation mechanism, competitive in anomaly detection
Other Models: Informer, iTransformer, TimeMixer, PatchTST, DLinear, TimeXer
IMPORTANT for Time Series: When using TSLib models, set data_preprocessing=["none"] and feature_extraction=["none"] since these models learn features internally.
3. Scikit-learn (Library: sklearn)
Applicable Data Type: Tabular Data
Description: General-purpose machine learning library
Anomaly Detection Models: IsolationForest, LocalOutlierFactor, OneClassSVM, EllipticEnvelope
For Classification Tasks (task_type: classification, multi_class_classification, multi_label_classification)
Scikit-learn (Library: sklearn)
Applicable Data Type: Tabular Data
Description: Supervised learning library with a wide range of classifiers
Available Models:
Linear Models: LogisticRegression, SGDClassifier, Perceptron, PassiveAggressiveClassifier
Tree-Based Models: DecisionTreeClassifier, RandomForestClassifier, ExtraTreesClassifier
Ensemble Methods: GradientBoostingClassifier, AdaBoostClassifier, VotingClassifier, StackingClassifier
Others: SVC, KNeighborsClassifier, GaussianNB, MLPClassifier
For Regression Tasks (task_type: regression)
Scikit-learn (Library: sklearn)
Applicable Data Type: Tabular Data
Description: Supervised learning library with regression algorithms
\end{lstlisting}
\begin{lstlisting}[style=promptstyle]
Available Models:
Linear Models: LinearRegression, Ridge, Lasso, ElasticNet, BayesianRidge
Tree-Based Models: DecisionTreeRegressor, RandomForestRegressor, ExtraTreesRegressor
Ensemble Methods: GradientBoostingRegressor, AdaBoostRegressor, HistGradientBoostingRegressor
Others: SVR, KNeighborsRegressor, MLPRegressor, GaussianProcessRegressor
For Forecasting/Prediction Tasks (task_type: forecasting)
TSLib (Library: tslib)
Applicable Data Type: Time Series Data
Description: Deep learning library specialized for time series forecasting
Available Models: Autoformer, Informer, TimesNet, iTransformer, TimeMixer, PatchTST, DLinear, FEDformer, TimeXer
Scikit-learn (Library: sklearn)
Applicable Data Type: Time Series Data (simple cases)
Description: Can be used for basic time series forecasting via supervised regression models
Available Models: LinearRegression, Ridge, and other standard regressors
\end{lstlisting}

\begin{lstlisting}[style=promptstyle]
VARIANT {variant_index}/{total_variants} REQUIREMENTS
Rules:
1. Each category (data_preprocessing, feature_extraction, anomaly_algorithm) must use EXACTLY ONE method
2. At least ONE category must use a DIFFERENT method from previous variants
3. Parameter-only changes are NOT allowed - must change method names
4. Diversity in preprocessing/feature_extraction: Some variants MAY use alternatives (e.g. ResNet/WideResNet patch features, PatchCore-style backbone). Do NOT use YOLOv5/YOLOv8 or object detection models for image anomaly detection.
\end{lstlisting}

\begin{lstlisting}[style=promptstyle]
You are a planner for a machine learning system. {intro}
{context}
{variant_guidance}
{sop_middle_block}
{efficiency_block}

PLAN STEPS
1. data_preprocessing: EXACTLY ONE method. You may use any preprocessing you deem appropriate (e.g. none, MinMaxScaler, StandardScaler, fill_median, or others). "none" is often enough when there are no special requirements. For time series with TSLib models: USE ["none"].
2. feature_extraction: EXACTLY ONE method. Image: PatchCore, CNN_backbone, pretrained CNN patch features (no YOLOv/object detection). Graph: graph_statistical_features. Tabular: PCA, statistical_features. Do NOT use PyGOD (pygod) for graphs. TSLib: USE ["none"].
3. anomaly_algorithm: EXACTLY ONE algorithm (match task_type: supervised/unsupervised/time_series). Use any algorithm you think suitable (e.g. from PyOD, sklearn, xgboost, tslib, darts, etc.).

Step format: step_id, agent_type, order, dependencies, methods (EXACTLY ONE per category), method_params

Output JSON (example with simple preprocessing/extraction):
{
  "metrics": ["accuracy", "precision", "recall", "f1_score"],
  "steps": [
    {"step_id": "step_1", "agent_type": "data_preprocessing", "order": 1, "dependencies": [], "methods": ["none"], "method_params": {"none": {"notes": "Skip preprocessing - use raw data"}}},
    {"step_id": "step_2", "agent_type": "feature_extraction", "order": 2, "dependencies": ["step_1"], "methods": ["none"], "method_params": {"none": {"notes": "Pass through"}}},
    {"step_id": "step_3", "agent_type": "anomaly_algorithm", "order": 3, "dependencies": ["step_2"], "methods": ["RandomForestClassifier"], "method_params": {"RandomForestClassifier": {"n_estimators": 100, "max_depth": 10}}}
  ]
}

METRICS
Include "metrics" field: classification/regression->["accuracy", "f1_score"], anomaly_detection->["precision", "recall", "f1_score"], forecasting->["mse", "rmse", "mae"]

NOTES
- methods: EXACTLY ONE per category (data_preprocessing, feature_extraction, anomaly_algorithm). You are free to choose any method; reference lists are for inspiration only.
- DO NOT use PyGOD (pygod) for graph data. Use graph_statistical_features + PyOD/sklearn or other alternatives.
- method_params: {"MethodName": {"param": value}} or {"none": {"notes": "..."}} for skip steps.
- If data_rows > 4,000 and using window-based features, set window_step_seconds >= 5.
- Keep model size moderate: avoid extremely high-dimensional hidden layers, feature vectors or too large n_estimators; prefer small/medium configurations for speed.

Output only JSON.
\end{lstlisting}
\subsection{IAD Code Generator Prompt}

\begin{lstlisting}[style=promptstyle]
You are a code generation expert. Generate a complete, executable Python script strictly following the plan, DO NOT modify, change, or use different values for these parameters.

IMPORTANT OUTPUT RULE:

Output ONLY Python code.
NO markdown, NO explanations outside code.
Start with import statements, end with the last line of code.

CRITICAL: The output must be raw Python source code only. Do NOT wrap it in Markdown code fences. Do NOT output lines like ```python or ``` anywhere.

The context_info is: {context_info}

The error_hint is: {error_hint}

======REQUIREMENTS (MANDATORY)
A) SCRIPT MUST BE COMPLETE & STANDALONE

Include ALL necessary imports and helper functions.
At the VERY BEGINNING of the script, include automatic package installation logic for uncommon packages.
Use try-except imports for uncommon packages; on ImportError, auto-install with pip then re-import.
Common packages (e.g., pandas, numpy, scikit-learn, scipy) may be imported directly without installation logic. Only add try-except + install for packages NOT in the standard scientific Python stack.

CRITICAL: Scikit-learn Version Compatibility
- For OneHotEncoder: Use 'sparse_output=False' (NOT 'sparse=False'). The 'sparse' parameter was deprecated and replaced with 'sparse_output' in scikit-learn >= 1.2.
- Example: OneHotEncoder(sparse_output=False, handle_unknown='ignore', drop=None)
- Always use 'sparse_output' instead of 'sparse' to ensure compatibility with modern scikit-learn versions.

B) PACKAGE INSTALLATION (CRITICAL)
Must implement install_package() and use subprocess + pip: subprocess.check_call([sys.executable, "-m", "pip", "install", package_name])
Uncommon packages (examples): skmultilearn, pyod, pygod, darts, imbalanced-learn, etc.
Pip name may differ from import name (e.g., scikit-multilearn vs skmultilearn).
PyGOD IMPORT RULE (CRITICAL)
If use pygod, import detectors ONLY from 'pygod.detector' (NOT 'pygod.models'). Example: from pygod.detector import DOMINANT

CRITICAL: NumPy Version Compatibility
If using PyTorch/ultralytics, check NumPy version at script start and downgrade if NumPy >= 2.0:

import numpy as np
if int(np.__version__.split('.')[0]) >= 2:
    import subprocess, sys, importlib
    subprocess.check_call([sys.executable, "-m", "pip", "install", "numpy<2.0", "--quiet"])
    importlib.reload(np)

Do this BEFORE importing PyTorch/ultralytics.

C) REQUIRED FUNCTIONS
Implement these functions (based on plan steps and methods/params):

Data Preprocessing: def preprocess_data(df: pd.DataFrame) -> pd.DataFrame

Feature Extraction: def extract_features(df: pd.DataFrame) -> pd.DataFrame
CRITICAL: When computing statistical features (mean, std, median, min, max, quantiles, etc.), MUST convert DataFrame to numpy array first (e.g., feature_matrix = df[columns].values) then use batch/vectorized operations (e.g., np.mean(feature_matrix, axis=1)). NEVER use row-by-row iteration (iterrows(), apply() with lambda) for statistical calculations.
For sliding window feature extraction: MUST create all window indices at once (e.g., start_indices = np.arange(0, n_samples - window_size + 1, window_step); end_indices = start_indices + window_size) then batch slice the feature matrix. NEVER loop through windows one by one to create slices. Stack all windows into 3D array [n_windows, window_size, n_features] then compute statistics along axis=1 using vectorized operations. For skew/kurtosis and delta features, compute across all features at once using vectorized numpy operations.

Modeling:
If anomaly_detection: def detect_anomalies(X) -> Tuple[np.ndarray, np.ndarray] returns (anomaly_labels, anomaly_scores)
Else (supervised tasks): def train_model(X, y) -> Any

CRITICAL API & DATA TYPE RULES:
1. PyGOD vs PyOD API Difference:
   - PyGOD models (e.g., DOMINANT): Use 'decision_score_' (singular), returns torch.Tensor
   - PyOD models: Use 'decision_scores_' (plural), returns numpy.ndarray
   - Example: anomaly_scores = model.decision_score_  # for PyGOD
   - Example: anomaly_scores = model.decision_scores_  # for PyOD

2. PyOD AutoEncoder Parameter Names (CRITICAL):
   - PyOD AutoEncoder uses 'hidden_neuron_list' (NOT 'hidden_neurons')
   - PyOD AutoEncoder uses 'epoch_num' (NOT 'epochs')
   - PyOD AutoEncoder uses 'lr' (NOT 'learning_rate')
   - PyOD AutoEncoder uses 'optimizer_name' (NOT 'optimizer')
   - PyOD AutoEncoder uses 'hidden_activation_name' (NOT 'activation' or 'hidden_activation')
   - Correct example:
     from pyod.models.auto_encoder import AutoEncoder
     model = AutoEncoder(
         hidden_neuron_list=[64, 32],  # NOT hidden_neurons
         epoch_num=100,                 # NOT epochs
         lr=0.001,                      # NOT learning_rate
         optimizer_name='adam',          # NOT optimizer
         hidden_activation_name='relu',  # NOT activation
         batch_size=32,
         contamination=0.1,
         preprocessing=True,
         device=None,
         random_state=42
     )
   - WRONG: AutoEncoder(hidden_neurons=[64, 32], epochs=100)  # These parameter names do NOT exist
   - Always check PyOD documentation for correct parameter names before using

3. PyOD VAE Parameter Names (CRITICAL):
   - PyOD VAE uses 'encoder_neuron_list' (NOT 'encoder_neurons')
   - PyOD VAE uses 'decoder_neuron_list' (NOT 'decoder_neurons')
   - PyOD VAE uses 'lr' (NOT 'learning_rate')
   - Correct example: VAE(encoder_neuron_list=[32, 16], decoder_neuron_list=[16, 32], lr=0.001, epoch_num=100, contamination=0.05)
   - WRONG: VAE(encoder_neurons=[32, 16], decoder_neurons=[16, 32], learning_rate=0.001)

4. PyOD Models WITHOUT random_state (CRITICAL):
   - COPOD, HBOS, ECOD do NOT support 'random_state' parameter - do NOT pass it
   - WRONG: COPOD(contamination=0.05, random_state=42)  # TypeError
   - CORRECT: COPOD(contamination=0.05, n_jobs=-1)
   - CORRECT: HBOS(n_bins=10, contamination=0.05)
   - CORRECT: ECOD(contamination=0.1)

4b. PyOD CBLOF beta parameter (CRITICAL):
   - CBLOF requires beta in range [1, 2147483647]. Do NOT use beta=0.1 or any value < 1.
   - WRONG: CBLOF(alpha=0.9, beta=0.1, ...)  # ValueError: beta is set to 0.1. Not in the range of (1, 2147483647)
   - CORRECT: CBLOF(alpha=0.9, beta=5, ...) or beta=1

5. PyOD ECOD contamination (CRITICAL):
   - ECOD 'contamination' MUST be a float (e.g., 0.1), NOT the string 'auto'
   - WRONG: ECOD(contamination='auto')  # AttributeError: 'str' object has no attribute 'eval'
   - CORRECT: ECOD(contamination=0.1)

6. PyTorch Tensor to NumPy Conversion (MANDATORY):
   - PyGOD/PyTorch models return torch.Tensor, but numpy functions (np.sum, np.mean, etc.) require numpy arrays
   - ALWAYS convert tensors before using numpy functions:
     if torch.is_tensor(anomaly_labels):
         anomaly_labels = anomaly_labels.cpu().numpy()
     if torch.is_tensor(anomaly_scores):
         anomaly_scores = anomaly_scores.cpu().numpy()
     # Or use: anomaly_labels = np.asarray(anomaly_labels)
   - Example: np.sum(anomaly_labels) requires anomaly_labels to be numpy array, not tensor

7. Data Type Consistency:
   - Ensure all arrays passed to sklearn functions have consistent dtypes (float32 or float64)
   - Convert PyTorch tensors to numpy before sklearn operations
   - Use .astype(np.float32) or .astype(np.float64) if dtype mismatches occur

{DATASET_SPECIFIC_SECTION}

E) TRAIN-TEST SPLIT RULES

CRITICAL: The dataloader already returns separated train_x, train_y, test_x, test_y. DO NOT re-split the data.

IMPORTANT RULES:
1. NEVER merge train and test data then re-split: The dataloader has already separated the data. Merging and re-splitting would:
   - Cause data leakage (test data may contain training samples)
   - Break the original train/test distribution
   - Lead to incorrect evaluation results

2. Use train and test data separately:
   - For preprocessing: Fit scalers/preprocessors on train_df, then transform test_df
   - For feature extraction: Fit extractors on train_df, then transform test_df
   - For training: Use processed train data only
   - For evaluation: Use processed test data only

3. The is_supervised flag determines how to handle labels:
   - is_supervised=True: Extract labels from train_df and test_df, use for supervised training/evaluation
   - is_supervised=False or None: Ignore labels (for unsupervised anomaly detection)

4. Label column identification:
   - All dataloaders return label column as 'anomaly_label' (check for 'anomaly_label', then 'Normal/Attack', then 'label', else use last column)
   - Extract labels from train_df and test_df separately: train_y = train_df['anomaly_label'], test_y = test_df['anomaly_label']

F) EVALUATION METRICS + JSON SAVE (MANDATORY)

Metrics computation and the exact evaluation_metrics dict format are defined in the DATA-MODALITY-SPECIFIC section below (graph / image / numeric / time_series). Follow that section for:
- How to compute metrics for {task_type} (per-class then macro for numeric; binary or forecasting for others)
- The required keys and types for evaluation_metrics
- Save path: result/metrics; filename: {metrics_timestamp}_{dataset_name}{plan_suffix}.json (use exact "{metrics_timestamp}" and "{plan_suffix}" from plan); json.dump(..., indent=2, ensure_ascii=False); print the saved file path.

G) RESULTS RETURN

main_pipeline must return: results_dict (single dict containing intermediate outputs and final evaluation_metrics).

{gpu_instructions}

=====TASK TYPE + PLAN (MUST FOLLOW)
Task Type: {task_type}
Is Supervised: {is_supervised}
Metrics to compute: {metrics}

PLAN STEPS (implement each step's method + method_params in order):
{plan_steps_json}

Generate the complete Python code now.
\end{lstlisting}